\documentclass[sigconf]{acmart}

\AtBeginDocument{%
  \providecommand\BibTeX{{%
    \normalfont B\kern-0.5em{\scshape i\kern-0.25em b}\kern-0.8em\TeX}}}

\copyrightyear{2023}
\acmYear{2023}
\setcopyright{rightsretained}
\acmConference[KDD '23]{Proceedings of the 29th ACM SIGKDD Conference on
Knowledge Discovery and Data Mining}{August 6--10, 2023}{Long Beach, CA,
USA}
\acmBooktitle{Proceedings of the 29th ACM SIGKDD Conference on Knowledge
Discovery and Data Mining (KDD '23), August 6--10, 2023, Long Beach, CA, USA}
\acmDOI{10.1145/3580305.3599360}
\acmISBN{979-8-4007-0103-0/23/08}

\acmPrice{15.00}

\usepackage{algorithm}
\usepackage{algorithmic}
\usepackage{subfigure}
\usepackage{multirow}
\usepackage[normalem]{ulem}
\useunder{\uline}{\ul}{}
\usepackage{wrapfig}
\usepackage{epstopdf}
\usepackage{xspace}
\usepackage{amsthm}
\usepackage{color}
\usepackage{hyperref}

\newcommand{\rulesep}{\unskip\ \vrule\ }

\newcommand{\method}{{\ttfamily DDLearn}\xspace}

\usepackage{enumitem}
\setlist{leftmargin=4mm}

\begin{document}

\title{Generalizable Low-Resource Activity Recognition with Diverse and Discriminative Representation Learning}

\author{Xin Qin}
\email{qinxin18b@ict.ac.cn}
\orcid{0000-0002-4717-4310}
\affiliation{%
  \institution{Beijing Key Lab. of Mobile Com., CAS}
  \country{China}
}

\author{Jindong Wang}
\authornote{Corresponding author: Jindong Wang.}
\email{jindong.wang@microsoft.com}
\orcid{}
\affiliation{%
  \institution{Microsoft Research Asia}
  \country{China}
}

\author{Shuo Ma}
\email{mashuo20g@ict.ac.cn}
\orcid{}
\affiliation{%
  \institution{Beijing Key Lab. of Mobile Com., CAS}
  \country{China}
}

\author{Wang Lu, Yongchun Zhu}
\email{{luwang, zhuyongchun18s}@ict.ac.cn}
\orcid{}
\affiliation{%
  \institution{Beijing Key Lab. of Mobile Com., CAS}
  \country{China}
}

\author{Xing Xie}
\email{xingx@microsoft.com}
\orcid{}
\affiliation{%
  \institution{Microsoft Research Asia}
  \country{China}
}

\author{Yiqiang Chen}
\email{yqchen@ict.ac.cn}
\orcid{}
\affiliation{%
  \institution{Beijing Key Lab. of Mobile Com., CAS}
  \country{China}
}

\renewcommand{\shortauthors}{Xin Qin et al.}

\begin{abstract}
Human activity recognition (HAR) is a time series classification task that focuses on identifying the motion patterns from human sensor readings. Adequate data is essential but a major bottleneck for training a generalizable HAR model, which assists customization and optimization of online web applications. However, it is costly in time and economy to collect large-scale labeled data in reality, i.e., the \emph{low-resource} challenge. Meanwhile, data collected from different persons have \emph{distribution shifts} due to different living habits, body shapes, age groups, etc. The low-resource and distribution shift challenges are detrimental to HAR when applying the trained model to new \emph{unseen} subjects. In this paper, we propose a novel approach called \textbf{D}iverse and \textbf{D}iscriminative representation \textbf{Learn}ing~(\method) for generalizable low-resource HAR. \method simultaneously considers diversity and discrimination learning. With the constructed self-supervised learning task, \method enlarges the data diversity and explores the latent activity properties. Then, we propose a diversity preservation module to preserve the diversity of learned features by enlarging the distribution divergence between the original and augmented domains. Meanwhile, \method also enhances semantic discrimination by learning discriminative representations with supervised contrastive learning. Extensive experiments on three public HAR datasets demonstrate that our method significantly outperforms state-of-art methods by an average accuracy improvement of \textbf{9.5\%} under the low-resource distribution shift scenarios, while being a generic, explainable, and flexible framework. Code is available at: \url{https://github.com/microsoft/robustlearn}.
\end{abstract}





\maketitle

\section{Introduction}

\label{sec-int}

Human Activity Recognition~(HAR) plays an indispensable role in ubiquitous computing. The goal of HAR is to build models by leveraging data such as the acceleration and rotation of angles recorded by inertial measurement units or other sensor devices to recognize users' activities.
With the development of machine learning and deep learning techniques, wearable sensor-based HAR has been applied to many real-life scenarios such as fatigue detection during the driving procedure~\cite{bulling2014tutorial}, fall detection of the elders~\cite{mauldin2018smartfall}, healthcare in daily life, and diagnosis of the Parkinson's disease~\cite{hammerla2015pd, chen2020fedhealth}, etc.
These applications show strong need for designing generalizable and accurate HAR algorithms in real-life scenarios.

\begin{figure}[t!]
	\centering
	\vspace{-.1in}
	\subfigure[Low-resource. Different sensor readings are collected from the same subject at different time.]{
		\includegraphics[scale=0.4]{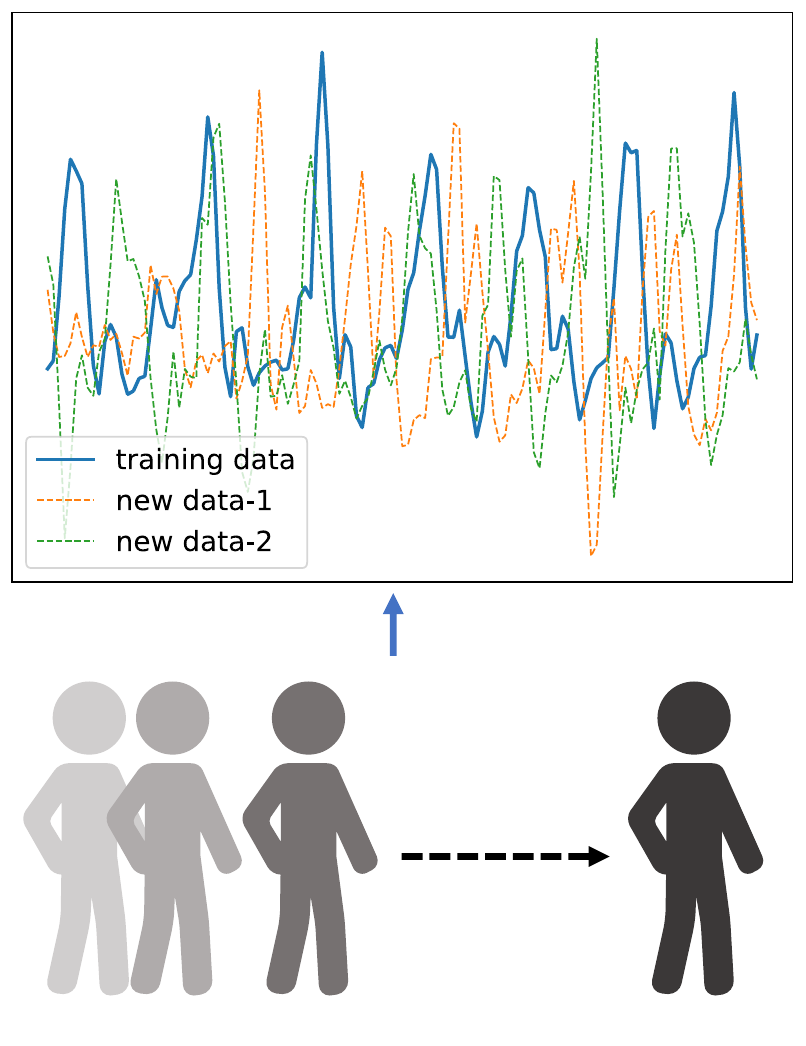}
		\label{fig-lowresource}}
	\subfigure[Distribution shift. Different sensor readings are collected from different subjects.]{
		\includegraphics[scale=0.4]{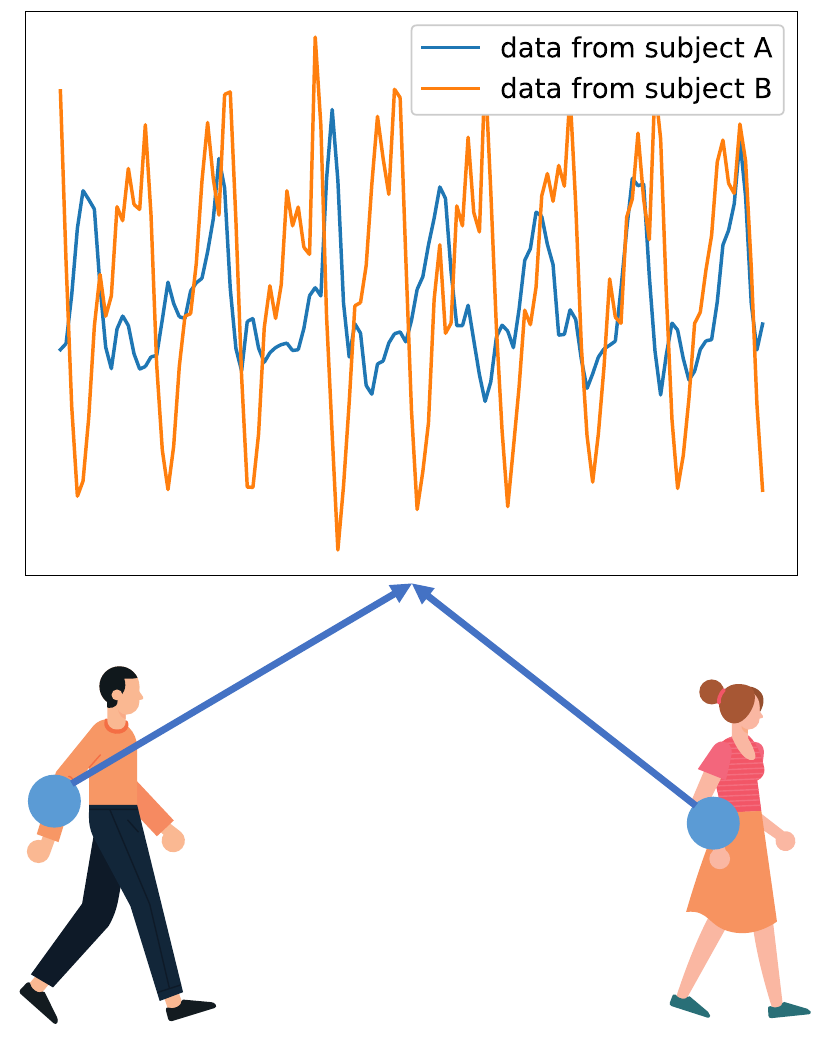}
		\label{fig-distribution}}
	\vspace{-.1in}
	\caption{Low-resource and distribution shift problems in HAR. Sensor readings are \textit{walking} activity in DSADS dataset.}
	\vspace{-.2in}
	\label{fig-intro}
\end{figure}

Despite the great progress, current HAR still faces two critical challenges that prevent us from building a generalizable model to perform well on \emph{unseen} data.
First, the \emph{low-resource} problem. 
While massive labeled data is indispensable for training powerful deep learning models, it is difficult to collect sufficient user data and annotate them in reality.
As shown in Figure~\ref{fig-lowresource}, the limited training data (the solid line) actually \emph{cannot} represent the diverse activity patterns that may change over time (the dashed lines).
Therefore, the \emph{low-resource} collected data fails to represent the rich and diverse patterns that can be used to learn a generalized model.
Second, the \emph{distribution shift} problem. 
From Figure~\ref{fig-distribution}, the data collected from different people have distribution discrepancies due to their different living habits, body shapes, age groups, etc.
Building models without considering the Non-IID (i.e., not identically and independently distributed) situation may result in great performance degradation since traditional machine learning often assumes that the training and test data are IID.
For example, a model trained on the data from existing patients may fail if applied to new patients with totally different body statuses.

How to solve the generalizable low-resource HAR problem?
Domain generalization (DG)~\cite{wang2021generalizing} is an emerging paradigm for solving the distribution shift problem between the source and target data.
Different from domain adaptation where the target data can be accessed in the training procedure, DG cannot access the target data, which is more challenging.
Unfortunately, existing DG algorithms cannot be used directly for our low-resource distribution shift problem since limited training data might fail to capture the diverse patterns which undermine the generalization abilities.

\begin{figure*}[t!]
  \centering
  \includegraphics[width=0.9\textwidth]{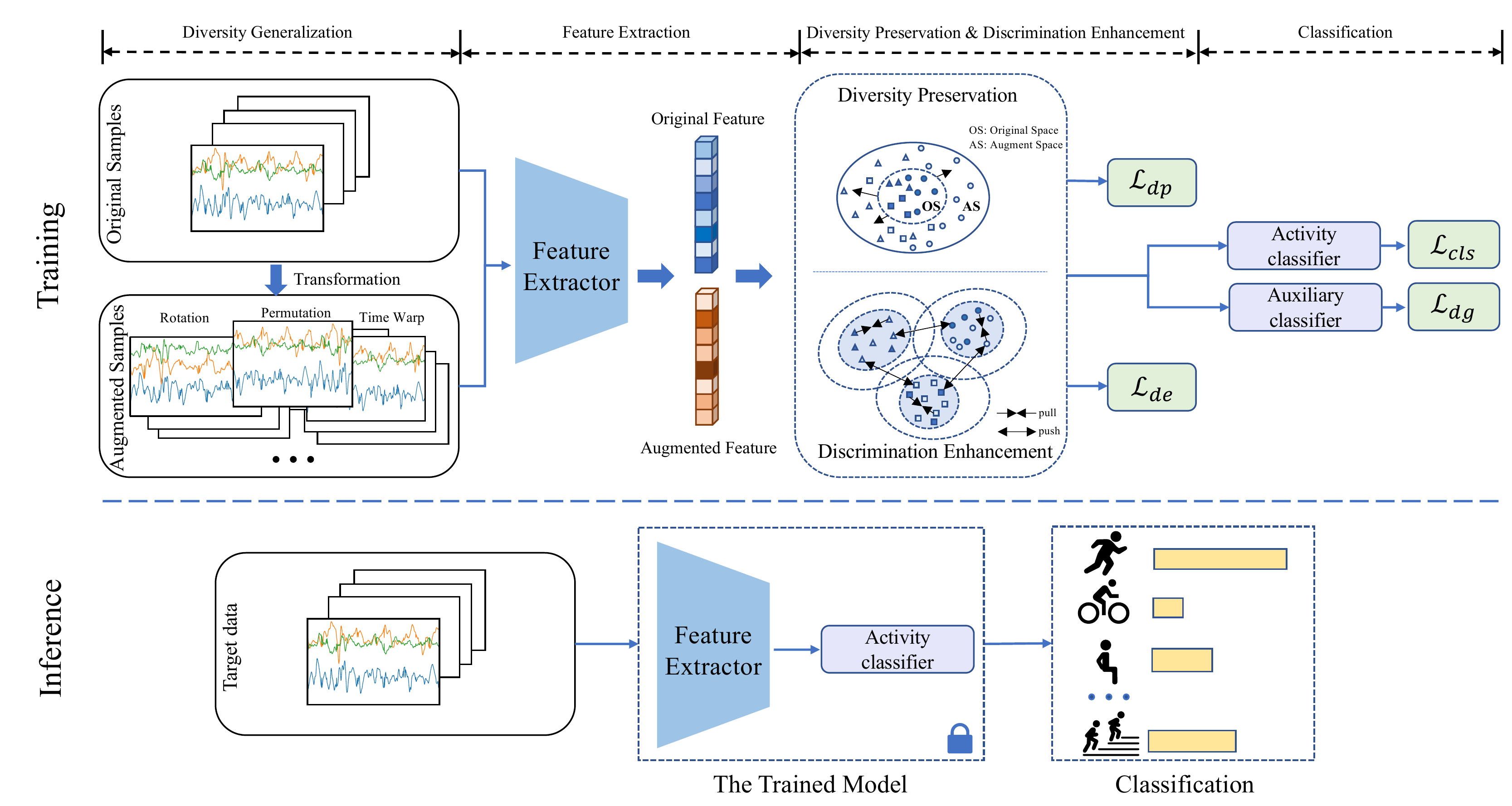}
  \vspace{-.1in}
  \caption{The training and inference procedures of \method.}
  \label{fig-method}
  \vspace{-.1in}
\end{figure*}

In this paper, we propose a novel approach named \emph{Diverse and Discriminative representation Learning}~(\method) for Generalizable Low-Resource HAR.
Our goal is to learn diverse and discriminative representations that can generalize to unseen target data with limited training samples.
We design a self-supervision task with particularly-designed sensor data transformations to solve the low-resource problem.
To further tackle the distribution shift problem, the learned representations should be more diverse and discriminative to ensure stronger generalization capability.
The main procedures of \method are illustrated in Figure~\ref{fig-method}: Diversity generalization, diversity preservation, and discrimination enhancement modules.
With data augmentation and the self-supervision auxiliary task, we expand the diversity of data space in semantic scope.
With diversity preservation, we preserve the diversity between the learned representations of the generated and original data, avoiding the degradation of diversity during the feature extraction procedure.
In order to achieve accurate activity recognition, the semantic-discriminative capability is also essential.
Thus, we further enhance the discriminating capability of representation learning.

Our main contributions are summarized as follows:
\begin{itemize}
    \item We aim to tackle a more challenging and realistic problem: \textit{generalizable low-resource} HAR. This brings two critical challenges: extremely limited training data and distribution shift.
    \item We propose a novel \method framework to solve this generalizable low-resource HAR problem by considering the diversity and discrimination in learning processes.
    \item To solve the low-resource challenge, we propose to use the self-supervision technique to design an auxiliary task to learn the latent motion properties and expand data space with diversity. We then propose a diversity preservation module.
    \item We propose a discrimination enhancement module for the semantic consistency and discrimination of activities by pulling together the intra-class representations and pushing away the inter-class pairs with supervised contrastive learning.
    \item We make comprehensive evaluations on three public sensor-based datasets, showing that \method effectively improves the HAR performance under low-resource domain generalization scenarios while remaining generic, explainable, and flexible. 
\end{itemize}

\section{Related Work}
\label{sec-related}
\paragraph{\textbf{Human Activity Recognition (HAR)}}
HAR focuses on making classification of different activities that happen in daily living. According to the type of data, HAR can mainly be divided into two categories: sensor-based and vision-based~\cite{dang2020sensor}. Vision-based HAR collects data with optical devices\cite{yang2016super} but may encounter some security problems, for example, sensitive information such as facial information and irrelevant people may be accidentally disclosed on camera. Sensor-based HAR captures the activity data through environment deployed sensors or wearable sensors. 
The wide usage of the IMU-deployed wearable sensors makes it more convenient and practical to record activity data in people's daily life\cite{jain2022collossl}. So we mainly focus on the wearable sensor-based HAR problem.

To solve sensor-based HAR, many machine learning-based methods~\cite{anguita2012human,qin2019cross} are proposed and with the fast development of deep learning techniques, the performance of HAR has been significantly improved\cite{li2021meta, hammerla2016deep}. Despite the great progress that has been achieved, most of these methods are based on the assumption of training data are sufficient and the training and testing data have independently identical distributions. However, low-resource and distribution shift of data are two realistic and long-standing problems hindering generalization performance for new unseen data in HAR.

\paragraph{\textbf{Domain Generalization (DG)}}

DG~\cite{wang2021generalizing} is an emerging technology of transfer learning scope~\cite{wu2018two} and is becoming popular in recent years. DG aims to learn a robust and generalizable model with one or several source domains that have different probability distributions to get a minimized error on an unknown target domain. Different from domain adaptation~\cite{wang2018deep,wilson2020multi, lu2021cross} assuming the availability of the target/test data in training, DG focuses on the scenario that the target data is absent for training, and the model is directly applied to the target data without re-training or fine-tuning. It is worth noting that although in traditional machine learning, the testing set also cannot participate in training, the learning process is based on the assumption that the data is IID, and DG considers the out-of-distribution~(OOD) problem. This is more challenging but closer to real-life applications. In the computer vision community, DG is a popular topic and many effective domain generalization methods are emerging. Existing DG methods can be categorized into three branches~\cite{wang2021generalizing}: Data manipulation~\cite{yue2019domain,qiao2020learning,lu2022semantic}, Representation learning~\cite{jin2021style,zunino2021explainable,ludomain,lu2022local} and Learning strategy~\cite{zhou2021domain,li2018learning,carlucci2019domain,lu2022out}. Despite DG thriving in the computer vision community, there still very few works are proposed for human activity recognition tasks~\cite{qin2022domain}. Recently, to the best of our knowledge, the first work to solve the DG problem for HAR is proposed~\cite{qian2021latent} by disentangling domain-agnostic and domain-specific features.
However, there are still few existing methods try to solve the generalizable low-resource human activity recognition problem.

\paragraph{\textbf{Self-Supervised Learning (SSL)}}
SSL~\cite{liu2021self,zhang2022empirical,hao2022learnable} is a popular technique that can help alleviate the dependence on annotated data by learning representations with the supervision of self-defined pseudo labels. A popular and effective branch in SSL is to design a different but relative auxiliary task to pre-train the feature extractor for latent representations which can improve the learning of the downstream HAR tasks~\cite{saeed2021sense}. 
Recently, contrastive self-supervised learning is applied to HAR to improve the performance. 
Based on the contrastive learning framework SimCLR~\cite{chen2020simple}, \cite{khaertdinov2021contrastive} makes a combination with a transformer-based encoder to solve sensor-based HAR, and \cite{tang2020exploring} modifies SimCLR with time-series transformation functions. 
\cite{qian2022makes} explores several important components of the existing contrastive learning algorithms when applying them to HAR tasks from both algorithmic-level and task-level aspects. 

The major difference between our approach and existing self-supervised learning-based HAR is that they are based on the framework of \emph{pre-training and fine-tuning} or \emph{supervised} training on the labeled target activity data, but don't consider the OOD problems. Most of them assume the training and testing data are IID, and do not consider the more realistic situation that the test data usually has distribution shifts with the training data and cannot be available during the training. As aforementioned, directly applying them to the unseen test data may get sharp performance degradation. Thus, it is an essential and urgent challenge to tackle, generalization-orient designed method is important for learning more general representations for generalizable low-resource HAR.

\section{Proposed Approach}
\label{sec-method}

A domain can be defined as a joint probability distribution $\mathbb{P}_{X,Y}$ on $\mathcal{X}\times\mathcal{Y}$, where $\mathcal{X}$ and $\mathcal{Y}$ denote the instance space and label space, respectively~\cite{ding2017deep, qian2021latent}.
We are given a labeled source domain $\mathcal{D}^s=\left\{(\mathbf{x_i},y_i)\right\}_{i=1}^{n_s}$ with $n_s$ instances as the training set.
Note that $n_s$ is much smaller than the sample size of a normal training set and it is not enough to train a robust model.
The goal is to learn a model on the existing subjects' data $f:\mathcal{X}\rightarrow\mathcal{Y}$ which can generalize well on an \emph{unseen} test domain~(i.e. new subjects' data) $\mathcal{D}^t=\left\{(\mathbf{x_i},y_i)\right\}_{i=1}^{n_t}$ which can only be accessed in inference.
All domains share the same feature and label spaces while having different probability distributions~(joint distribution shift), i.e. $\mathcal{X}^s=\mathcal{X}^t, \mathcal{Y}^s=\mathcal{Y}^t, {P}^{s}(\mathbf{x_i},y_i)\neq{P}^{t}(\mathbf{x_i},y_i)$.
Note that the generalizable low-resource setting is more challenging than the conventional setting due to the small training data size and distribution shifts.

\subsection{Motivation and Overall Framework}
Two main issues, i.e., low-resource and distribution shift, seriously impede the capability of machine learning for HAR.
However, it is difficult and impractical to collect a large amount of data.
For instance, the elderly who have difficulty moving or children who have difficulty following the instruments do not support massive data collection.
Meanwhile, sensor readings can be easily affected by diverse personalities, leading to different data distributions between different users even if they perform the same activities with the same kind of sensors.
Enough data is important for training a good deep learning model~\cite{liu2021self}, especially when the test data has different distribution from the training data.
Thus, given limited training data, it is essential and intuitive to utilize data augmentation to explore the invariant properties of data by \emph{expanding} and completing the data space within the semantic scope~\cite{um2017data}.
In addition to data space expansion, it is also important to get \emph{diverse} representations that have strong generalization capability and keep the semantic discrimination capability for activity recognition.

In this paper, we propose a novel approach called \textbf{D}iverse and \textbf{D}iscriminative representation \textbf{Learn}ing~(\textbf{\method}) for generalizable low-resource HAR.
The core of \method is to learn a model which has strong generalization capability to OOD data with limited training instances.
As shown in Figure~\ref{fig-method}, \method~learns diversity both in data and feature space to overcome the limitation of low-resource training data.
Specifically, \method learns to generate diversity by data augmentation and exploring the invariant properties in the latent space by injecting the self-defined prior knowledge of augmentations in a self-supervised auxiliary task.
To learn generalizable representations in feature space, we propose a diversity preservation module and a discrimination enhancement module to preserve the representation diversity for accurate recognition. 

The overall learning objective of \method can be formulated as:
\begin{equation}
    \label{eq-all}
    \mathcal{L} = \mathcal{L}_{cls} + \lambda \mathcal{L}_{dg} + \beta \mathcal{L}_{dp} + \gamma \mathcal{L}_{de},
\end{equation}
where $\mathcal{L}_{cls}$ is the activity classification loss, $\mathcal{L}_{dg}$ is the loss of diversity generation, $\mathcal{L}_{dp}$ is the loss for diversity preservation, and $\mathcal{L}_{de}$ is the loss for discrimination enhancement learning. $\lambda$, $\beta$ and $\gamma$ are trade-off hyper-parameters.
In the following sections, we will elaborate on these learning modules.

\subsection{Diversity Generation}

In the low-resource scenario, it is hard to learn a generalized model due to the limited representation capabilities from limited data.
The distribution shift between the test and training data worsens the performance.
Therefore, we propose to generate diversity from both the data space and representation space by expanding data space with data augmentation and exploring the generalized and invariant properties by self-supervised auxiliary tasks.

Wearable sensor data differs from images because of its temporal property and motion patterns, thus the augmentation techniques should be well designed.
Specifically, we employ seven data transformation techniques~\cite{um2017data} to get the augmented data.
These data transformations are: \emph{rotation, permutation, time-warping, scaling scale, magnitude warping, jittering,} and \emph{random sampling}.~\cite{saeed2019multi} verified these transformations are feasible and practical for HAR. 
Details of these transformations are as follows.

\begin{itemize}
    \item{Rotation}: Rotate the data with an arbitrary angle that is sampled randomly from a uniform distribution. This transformation can simulate the various orientations of the sensors placed on different body locations when performing the same activity~(i.e. different sensor orientations with the same label).
    \item{Permutation}: Slice a window of data into \textit{N} segments and randomly permute these segments to form a new window. This transformation may help to explore the permutation invariant properties in learning.
    \item{Time-Warping}: Perturb the location in the temporal dimension by smoothly and locally distorting the time intervals between samples. This may broaden the local diversity in temporal.
    \item{Scaling}: Scale the magnitude of a window-length data by multiplying with a random scalar. This can enlarge the magnitude diversity of the entire samples with constant noise.
    \item{Magnitude Warping}: Different from scaling, it warps the magnitude of a window-length data with a smooth scalar around 1 which can add smoothly-varying noise to samples.
    \item{Jittering}: This applies different noise to samples. It may also enlarge the diversity of data magnitude and push the model to be more robust against the multiplicative and additive noise.
    \item{Random Sampling}: This transformation is similar to time-warping transformation, but it only uses subsamples for interpolation.
    
\end{itemize}

Motivated by existing work \cite{saeed2019multi} that utilized sensor transformations to design a multi-task self-supervised network, we design a self-supervised auxiliary task with the original and seven types of transformed data.
More differently, we design one multi-class classification task to reduce the complexity of the model.
Thus, we have two tasks: The original activity recognition task (i.e., $\mathcal{L}_{cls}$) and the self-supervised auxiliary task.

To explore the underlying properties of the generated sensor signals, the learning objective of the self-supervised auxiliary task is to classify which kind of transformation the input data belongs to~(the original data is also regarded as a category here, i.e., 8 categories).
The label space of the auxiliary task is denoted as $\mathcal{Y}_p$.
We adopt the standard cross-entropy loss for the self-supervision task to learn the self-supervision model $f_{s} : \mathcal{X}_{all}\rightarrow\mathcal{Y}_{p}$, where $\mathcal{X}_{all}=\mathcal{X}_{aug}\cup\mathcal{X}_{ori}$, formulated as:
\begin{equation}
    \label{dg}
    \mathcal{L}_{dg} (f_s; \mathcal{X}_{all}, \mathcal{Y}_{p}) = -\mathbb{E}_{(x, y_p)\in \mathcal{X}_{all}\times \mathcal{Y}_{p}} \sum_{k=1}^K y_p \log \delta_k(f_s(x)),
\end{equation}
where $K$ is the number of classes in the self-supervision task, $y_p$ is the ground-truth class label of the self-supervision task, $\delta_k(f_s(x))$ is the predicted probability and $\delta$ is the softmax function.

\subsection{Diversity Preservation}
\label{sec-dp}
After diversity generation in the data space, we extract general features of the augmentation and original data with CNN framework and learn the task-oriented features with fully-connected layers.
During the feature learning procedure, the representation space of the same category may be narrowed and features gather closer under the effect of the classification objective.
This may result in the reduction of diversity in data augmentation. Hence, it is important to preserve the \emph{diversity} of the learned features, i.e., to make the augmented features distinguished from the original ones such that the feature space is expanded.
To this end, we focus on enlarging the distance between the original and the augmentation feature spaces.
Note that, with the regularization of the semantic discrimination enhancement (details will be introduced in the next section) and activity classification loss, the distance can be held in a moderated range. For easy optimization, maximizing the domain distance is identical to minimizing the following:
\begin{equation}
    \mathcal{L}_{dp} = - dist(\mathcal{D}_{ori}, \mathcal{D}_{aug}).
    \label{equ-dp}
\end{equation}

According to~\cite{ben2006analysis, ganin2016domain}, the distribution divergence of two domains $\mathcal{D}_{ori}$ and $\mathcal{D}_{aug}$ can be defined as Definition~\ref{def-H}:

\begin{definition}[Distribution distance]
\label{def-H}
For data ${X}$ and domains $\mathcal{D}_{ori}$ and $\mathcal{D}_{aug}$ over $X$, the $\mathcal{H}$-divergence of the binary classifier set $\mathcal{H}=\left\{\eta \right\}$ between these two domains is
\begin{equation}
d_{\mathcal{H}}\left(\mathcal{D}_{\mathrm{ori}}^{X}, \mathcal{D}_{\mathrm{aug}}^{X}\right)=2 \sup _{\eta \in \mathcal{H}}\left|\operatorname{Pr}_{\mathbf{x} \sim \mathcal{D}_{\mathrm{ori}}^{X}}[\eta(\mathbf{x})=1]-\operatorname{Pr}_{\mathbf{x} \sim \mathcal{D}_{\mathrm{aug}}^{X}}[\eta(\mathbf{x})=1]\right|,
\end{equation}
where $Pr$ denotes the classification prediction probability.
\end{definition}

Thus, the distribution between two domains can be bridged by the Proxy
$\mathcal{A}-{distance}$~\cite{ben2006analysis}, which is defined as the error of building a linear binary classifier to discriminate two domains.
Denote the domain discrimination error as $\epsilon$ with a classifier $\eta$, then, the $\mathcal{A}-{distance}$ can be defined as:
\begin{equation}
    \label{adist}
    d_{\mathcal{A}(\mathcal{D}_{ori}, \mathcal{D}_{aug})} = 2(1-2\epsilon(\eta)).
\end{equation}

We aim to \emph{maximize} the divergence between the original and augmentation data, which is identical to minimizing the error of the binary classifier.
Thus, we propose to utilize a domain discriminator with a binary classifier to discriminate the original and augmentation domains.
By minimizing the domain discriminator error, the distance between two domains can be enlarged and these two domains can be discriminated. So the diversity preservation objective is to minimize the classification loss of the domain discriminator, and we find this discriminator does not need to be well trained, random conditions can get a good effect.

It can help preserve the features of two domains from overlapping too much to lose diversity.
Note that \method is not limited to a specific metric such as minimizing the loss of domain discriminator, we can easily utilize other distance metrics and maximize them as alternatives to enlarge the distribution of the original and augmentation space such as Maximum Mean Discrepancy~(MMD)~\cite{gretton2012kernel}, and Kullback-Leibler~(KL) divergence~\cite{kullback1997information} and so on (refer to Sec.~\ref{sec-exp-exten}).

\vspace{-2mm}
\subsection{Discrimination Enhancement}
To achieve accurate activity recognition, it is important to enhance the \emph{semantic discrimination} of representations.
We aim to enlarge the inter-class distance~(i.e., the distance between samples from different classes) and reduce the intra-class distance (i.e. the distance between samples from the same class) to achieve semantic discrimination enhancement.

We adopt the supervised contrastive loss to the original and augmented features~\cite{khosla2020supervised, wang2021learning} to enhance the activity semantic discrimination, and randomly regard each sample not only the augmentations as anchors.
Supervised contrastive learning~\cite{khosla2020supervised} first makes data augmentations with two random augmentations or called \textit{views} (e.g. rotation, permutation). It randomly regards an augmented sample as the anchor, then the positive samples are the other samples that belong to the same class as the anchor's, and others are regarded as negative samples. Due to the presence of labels, supervised contrastive learning can help achieve the aforementioned intra-class pulling and iter-class pushing.
Therefore, the enhancement loss is computed as:
\begin{equation}
    \label{de}
    \mathcal{L}_{de} = -\sum_{i \in \emph{I}} \frac{1}{\left|P(i)\right|} \sum_{p\in P(i)}{\log \frac{\exp(z_i \cdot z_p /\tau)}{{\sum_{a\in A(i)} \exp(z_i \cdot z^a / \tau)}}},
\end{equation}
where $I$ is the index set of the original and the augmented representations and index $i$ is the \emph{anchor}. $A(i) \equiv I \backslash {i}$, $P(i) \equiv \left\{p \in A(i) : \tilde{y_p} = \tilde{y_i}\right\}$ is the set of indexes of the representations which is in the same activity class with the $i_{th}$ representation, $z_p$ is the positive representation, and $\tau \in \mathbb{R}^+$ is the scalar temperature.

By applying contrastive loss on original and augmented representations, representations of the same class are pulled closer and representations of different classes are pushed further away from each other. Thus, the semantic discrimination is enhanced to achieve accurate activity recognition.

\subsection{\method for Low-Resource HAR}

We introduce the training and inference of \method for low-resource generalizable HAR.
As shown in Algo.~\ref{algo-method}, we first conduct data augmentation with the aforementioned seven transformations to get the augmented data.
The original and augmented data are concatenated into mini-batches with the $1:1$ ratio as input and then fed into the network. Then, representations for both the original and the augmented data are learned by the feature extractor.
With the diversity preservation module, the distance between the original and augmented space is enlarged to avoid their fusion. By discrimination enhancement, intra-class features are pulled closer and inter-class features are pushed away, thus the semantic discrimination is enhanced. Subsequently, all features are fed to the main activity classifier and auxiliary augmentation classifier.

As for inference, we directly apply the trained model to the unseen test data.
Without data augmentation and fine-tuning, only the original test data are fed into the network to extract features and only apply the activity classifier to them for activity recognition.

\begin{algorithm}[htbp]
	\caption{\method for low-resource activity recognition}
	\label{algo-method}

	\renewcommand{\algorithmicrequire}{\textbf{Input:}} 
	\renewcommand{\algorithmicensure}{\textbf{Output:}}
    \raggedright
    \underline{\textbf{\emph{Training:}}}
	\begin{algorithmic}[1]
		\REQUIRE
		The training domain $\mathcal{D}^{s}$, and hyper-parameters $\lambda,\beta$, $\gamma$.\\
		\ENSURE
		The trained model $\mathcal{M}$.\\
		\STATE Randomly initialize the model parameters $\theta$;\\
		\STATE Conduct data augmentation with data transformation techniques and get the augmented data.
		\WHILE{not converge}{
		    \STATE Sample a mini-batch $\mathcal{B}=\{\mathcal{B}^{ori}, \mathcal{B}^{aug}\}$ from the original and augmented data and concat them as $\mathbf{x}_{all}$;\\
		    \STATE Extract features $f_e(\mathbf{x}_{all})$ by the feature extractor $f_e$;\\
		    \STATE Learning diversity preservation with Eq.~\eqref{equ-dp} and get $\mathcal{L}_{dp}$;\\
		    \STATE Learning discrimination enhancement via Eq.~\eqref{de} ($\mathcal{L}_{de}$);\\
            \STATE Learning the auxiliary task classifier and calculate the self-supervised task loss $\mathcal{L}_{dg}$ according to Eq.~\eqref{dg};\\
            \STATE Get classification loss $\mathcal{L}_{cls}$ for the main activity classifier;
		    \STATE Calculate the total loss of \method according to Eq.~\eqref{eq-all};
		    \STATE Update the model parameter using Adam optimizer.\\}
		\ENDWHILE\\
		\end{algorithmic}
		\raggedright
		\underline{\textbf{\emph{Inference:}}}
		\begin{algorithmic}[1]
		\REQUIRE
		The trained model $\mathcal{M}$, target domain data $\mathcal{D}^t$.\\
		\ENSURE
		Classification results on the test domain.\\
		\FOR{$(x,y) \in \mathcal{D}^t$}
		\STATE Get the predict label $\hat{y}=\mathcal{M}(x)$;
		\ENDFOR
		\STATE Calculate the classification accuracy.
		\RETURN Classification results on target HAR data.
	    \end{algorithmic}
\end{algorithm}

\section{Experiments}
\label{sec-exp}
We evaluate \method via extensive experiments on the low-resource generalizable activity recognition problems to investigate the following research questions (RQs):
\begin{itemize}
    \item \textbf{RQ1 (Accuracy):} What is the effectiveness of \method in public datasets?
    \item \textbf{RQ2 (Robustness):} Is \method robust in different levels of low-resource settings, i.e., different sizes of training data?
    \item \textbf{RQ3 (Interpretability):} What are the contributions of each component in \method and how to interpret their importance?
    \item \textbf{RQ4 (Case study):} How does the \method improve the performance on each activity?
    \item \textbf{RQ5 (Compatibility):} What about the compatibility of \method or can it be a flexible framework?
\end{itemize}

\subsection{Setup}
\label{subsec-dataset}

\paragraph{Datasets}
We adopt three public activity recognition datasets.
(1) \textbf{DSADS}.
UCI Daily and Sports Data Set~\cite{barshan2014recognizing} collects data from 8 subjects around 1.14M samples. Three kinds of body-worn sensor units including triaxial accelerometer, triaxial gyroscope, and triaxial magnetometer are worn on 5 body positions of each subject: torso, right arm, left arm, right leg, and left leg. It consists of 19 activities. The total signal duration is 5 minutes for each activity of each subject.
(2) \textbf{PAMAP2}.
PAMAP2 Physical Activity Monitoring dataset~\cite{reiss2012introducing} consists of data collected from 9 subjects wearing 3 inertial measurement units (IMU) on hand, chest, ankle and a heart rate monitor~(HRM),  around 2.84M samples. We utilize the data collected with the IMU including triaxial accelerometer~(with the scale of $\pm16g$ as the official recommendation), gyroscope and magnetometer data. We utilize 8 common activities from 8 subjects for evaluation: lying, sitting, standing, walking, ascending stairs, descending stairs, vacuum cleaning, and ironing.
(3) \textbf{USC-HAD}.
USC Human Activity Dataset~\cite{zhang2012usc} collects data from 14 subjects with a motion mode packed into a mobile phone and attached to the front right hip of subjects, around 2.81M samples. During data collection, triaxial accelerometer and triaxial gyroscope sensor data are captured. Each subject performs 12 activities in their own styles.
\textbf{Detailed dataset information is in Appendix~\ref{app-preprocess}.}

\paragraph{Comparison methods}
We compare \method with Empirical Risk Minimization~(ERM)~\cite{vapnik1992principles} and recent OOD/SSL methods:
\begin{itemize}
    \item ERM~\cite{vapnik1992principles}: the baseline for domain generalization. 
    \item SimCLR~\cite{chen2020simple}:  generates positive samples with data augmentation and performs contrastive learning.
    \item Mixup~\cite{xu2020adversarial}: linear interpolations on random pairs of samples.
    \item MLDG~\cite{li2018learning}: meta-learns for domain generalization. 
    \item RSC~\cite{huang2020self}: discards the dominant activations on the training data.
    \item AND-Mask~\cite{parascandolo2021learning}: masks gradients via `AND' for invariances.
    \item Fish~\cite{shi2021gradient}: matches the gradient for domain generalization.
\end{itemize}

\subsection{Implementation Details}

\subsubsection{Data Pre-processing}

To construct the domain-generalized activity recognition scenario, we divide subjects into several groups for leave-one-out-validation, i.e. we regard one group of subjects' data as the target domain and the remained as source domain data, each can be regarded as a task. We divide 8 subjects into 4 groups for DSADS and PAMAP2 and divide 14 subjects into 5 groups where each of group 0-3 consists of 3 subjects and the last consists of 2 subjects. We further split each group's data into training, validation, and test data with a ratio of 6:2:2, and select the best model on the source validation set, meanwhile, we can further make a comparison with the ideal situation that straightly training on the target, which is a less practical situation for real-life applications.

We randomly sample \textbf{20\%-100\%} samples from the training data with a step of 20\% as the training set to evaluate the influence of the training data size, and simulate the low-resource setting. For testing, we evaluate the trained model on the test set of the target domain. 
\textbf{Details of using sliding window for pre-processing each sample are in Appendix~\ref{app-preprocess}.}

\begin{table*}[t!]{

\caption{Classification accuracy (\%) ($\pm$ standard deviation) on three public datasets in low-resource setting with only \textbf{20\%} of the training data. The best and the second-best results are marked with \textbf{bold} and \underline{underline}, respectively.}
\label{tab-acc-02data}
\begin{tabular}{lc|cccccccc}
\toprule
                         & Tar & ERM~\cite{vapnik1992principles}              & Mixup~\cite{xu2020adversarial}                   & Mldg~\cite{li2018learning}              & RSC~\cite{huang2020self}                     & AND-Mask~\cite{parascandolo2021learning}                & SimCLR~\cite{chen2020simple}             &Fish~\cite{shi2021gradient}       & Ours                       \\ \midrule
\multirow{5}{*}{\rotatebox{90}{DSADS}}   & T0  & 56.70 ($\pm$2.65) & {\ul 74.77 ($\pm1.76$)} & 59.45 ($\pm2.92$) & 54.32 ($\pm2.19$)       & 57.18 ($\pm2.91$)       &  72.48 ($\pm3.18$)      &  55.06 ($\pm1.60$)  & \textbf{87.88 ($\pm1.92$)} \\
                         & T1  & 61.05 ($\pm7.94$) & 75.78 ($\pm3.95$) & 64.07 ($\pm3.67$) & 63.62 ($\pm10.56$)      & 61.24 ($\pm1.35$)       & {\ul 76.61 ($\pm2.56$)}   &  62.28 ($\pm3.13$)   & \textbf{88.80 ($\pm1.11$)} \\
                         & T2  & 61.31 ($\pm7.04$) & 74.18 ($\pm4.36$) & 66.78 ($\pm1.15$) & 66.48 ($\pm1.80$)       & 66.67 ($\pm2.70$)       & {\ul 78.25 ($\pm0.92$)}   &  68.15 ($\pm1.60$)  & \textbf{89.21 ($\pm1.23$)} \\
                         & T3  & 66.48 ($\pm5.69$) & 75.85 ($\pm3.45$) & 69.16 ($\pm0.62$) & 64.29 ($\pm3.37$)       & 67.04 ($\pm3.72$)       & {\ul 76.49 ($\pm0.91$)}  &  68.83 ($\pm3.83$)   & \textbf{85.63 ($\pm1.13$)} \\
                         & Avg & 61.38 ($\pm3.82$) &  75.15 ($\pm2.36$) & 64.86 ($\pm1.24$) & 62.18 ($\pm4.32$)       & 63.03 ($\pm0.84$)       & {\ul75.96 ($\pm1.25$)}  & 63.58 ($\pm0.37$)   & \textbf{87.88 ($\pm0.82$)} \\ \midrule
\multirow{5}{*}{\rotatebox{90}{PAMAP2}}  & T0  & 48.18 ($\pm1.03$) & 57.81 ($\pm0.55$) & 56.64 ($\pm3.08$) & 55.99 ($\pm1.29$)       & 52.60 ($\pm5.79$)       &  {\ul63.28 ($\pm3.33$)}      &  54.04 ($\pm4.31$) & \textbf{75.55 ($\pm0.79$)} \\
                         & T1  & 79.95 ($\pm5.49$) & 81.51 ($\pm3.94$)       & 81.12 ($\pm1.87$) & 83.08 ($\pm2.42$)       & 81.77 ($\pm5.35$)       & 81.25 ($\pm1.59$)  &   {\ul 85.16 ($\pm1.39$)} & \textbf{90.07 ($\pm2.40$)} \\
                         & T2  & 74.35 ($\pm3.53$) & 77.34 ($\pm3.33$)       & 76.70 ($\pm2.86$) & 78.65 ($\pm3.99$)       & {\ul 80.08 ($\pm3.04$)} &   78.65 ($\pm1.87$)    &  79.69 ($\pm4.00$)  & \textbf{85.51 ($\pm0.76$)} \\
                         & T3  & 67.32 ($\pm7.78$) & 70.31 ($\pm5.64$)       & 71.87 ($\pm2.49$) & 68.10 ($\pm6.27$)       & 67.19 ($\pm5.75$)       &71.09 ($\pm1.99$)  & {\ul 72.53 ($\pm0.49$)} & \textbf{80.67 ($\pm1.78$)} \\ 
                         & Avg & 67.45 ($\pm2.09$) & 71.74 ($\pm1.37$)       & 71.58 ($\pm0.92$) & 71.45 ($\pm2.55$)       & 70.41 ($\pm3.60$)       & {\ul73.57 ($\pm1.21$)}  &  72.85 ($\pm0.37$) & \textbf{82.95 ($\pm0.60$)} \\ \midrule
\multirow{6}{*}{\rotatebox{90}{USC-HAD}} & T0  & 69.79 ($\pm3.34$) & 68.66 ($\pm4.67$)       & 67.45 ($\pm1.18$) & {\ul 75.69 ($\pm4.28$)} & 70.57 ($\pm4.06$)       &   69.36 ($\pm2.34$)     &   73.70 ($\pm3.97$)& \textbf{79.06 ($\pm2.11$)} \\
                         & T1  & 67.88 ($\pm1.00$) & 68.75 ($\pm1.29$)       & 64.32 ($\pm2.05$) & {\ul 72.40 ($\pm2.88$)} & 68.84 ($\pm0.98$)       & 66.62 ($\pm1.44$)  & 72.05 ($\pm2.93$)   & \textbf{80.15 ($\pm1.11$)} \\
                         & T2  & 67.62($\pm2.18$)  & 71.79 ($\pm0.65$)       & 63.80 ($\pm1.18$) & 72.83 ($\pm3.62$) & 69.18 ($\pm3.10$)       &  {\ul 76.04 ($\pm1.61$)}   &  69.10 ($\pm2.93$)   & \textbf{80.81 ($\pm0.74$)} \\
                         & T3  & 59.64 ($\pm6.26$) & 61.29 ($\pm3.90$)       & 56.86 ($\pm6.80$) & {\ul 63.19 ($\pm5.30$)} & 61.89 ($\pm4.48$)       & 61.24 ($\pm1.06$)    & 58.51 ($\pm3.66$)  & \textbf{70.93 ($\pm1.87$)} \\
                         & T4  & 59.46 ($\pm1.28$) & 65.63 ($\pm4.55$)       & 58.16 ($\pm1.05$) & {\ul 66.75 ($\pm3.25$)} & 59.81 ($\pm3.20$)       & 62.85 ($\pm2.17$)    & 63.72 ($\pm8.31$)  & \textbf{75.87 ($\pm2.99$)} \\ 
                         & Avg & 64.88 ($\pm2.49$) & 67.22 ($\pm2.41$)       & 62.12 ($\pm1.77$) & {\ul 70.17 ($\pm3.51$)} & 66.06 ($\pm2.54$)       & 67.22 ($\pm0.39$)    & 67.42 ($\pm3.91$)  & \textbf{77.36 ($\pm0.99$)} \\ \hline
                         \multicolumn{2}{c|}{Avg all}   & 64.57             & 71.37                   & 66.19             & 67.93                   & 66.50                   &  \underline{72.25}        &   67.95       & \textbf{82.73}                      \\ \hline
\end{tabular}}
\end{table*}

\subsubsection{Network Architecture and Training}
We directly follow \cite{qian2022makes} to reproduce SimCLR, since \cite{qian2022makes} has employed SimCLR to domain generalization HAR tasks with sensor data transformations.
Other methods are popular domain generalization methods and we apply them to HAR with the same backbone network (also a CNNs architecture) for a fair comparison following DomainBed~\cite{gulrajani2020search}.
The feature extractor includes conv2d layers (two conv2d layers with kernel of $(1, 9)$ for DSADS and PAMAP2, 3 conv2d layers with kernel size of $(1, 6)$ for USC-HAD), each along with a ReLU and maxpool2d operation and then connect with a fully connected layer for higher layer feature extraction.
The output feature dimension is 64 for DSADS and PAMAP2 and 128 for USC-HAD.
We utilize a fully connected layer as the classifier which uses the features as input and outputs the class-number-dimension logits.
By utilizing a softmax operation, we can get the prediction probability of each class that adds up to 1.
We set the learning rate to 0.0008 and use batch size 64 for the original data and 64 for the augmented data in our method.
Adam optimizer is utilized to optimize the training process.
\method is implemented with PyTorch and trained on GTX 3090.
We repeat each experiment 3 times with 3 different random seeds and report the mean and standard deviation.

\vspace{-2mm}
\subsection{Overall Performance~(RQ1)}
\label{res}

As the aforementioned division of subjects' data, each group is regarded as the testing target domain in turn and denoted with $\left\{T0, T1,...\right\}$ as their indexes, and the reminders together are regarded as the source domain.
We make the experimental evaluation of low-resource regime that only use 20\% of the training data.
The classification accuracy on three public datasets is shown in Table~\ref{tab-acc-02data}.
From the results, we observe that \method significantly improves the classification accuracy by \textbf{11.93}\% on the DSADS dataset, \textbf{9.38}\% on the PAMAP2 dataset, and \textbf{7.19}\% on the USC-HAD dataset, respectively, i.e., an average improvement of \textbf{9.5}\% on three datasets. This indicates that our method can achieve accurate activity recognition and makes good generalization to unseen new data without external fine-tuning, thus can solve the generalizable low-resource HAR challenges.
Besides, we observe that when facing more challenging tasks in which other methods get degraded accuracy such as the first task of PAMAP2 dataset, our method still can significantly improve the performance with \textbf{12.27}\%.
This indicates that our method is more robust to deal with difficult tasks.
Compared with directly applying the contrastive self-supervised learning method SimCLR to generalizable HAR, the performance of the proposed method is better, which indicates the diverse and discriminative representation learning is effective for generalization.
The standard deviation of \method is relatively small compared with the others when experimenting with three trials, implying that our method is stable.
We also make an experiment that pre-training a model on the source training data and then fine-tuning it on the target data.
The accuracy results can be significantly improved after fine-tuning.
However, it is less practical in the real-life application where we want to straightly apply the trained model to new subjects' data.
The impossibility of accessing increasing emerging new data during the training process and the lack of labels for new data 
hinder the application of fine-tuning from real-life HAR applications.  
\begin{table}[t!]
\centering
\caption{Classification accuracy (\%) on three public datasets with different percentages (\%) of training data.}
\label{tb-differ-data-rate}
\resizebox{.48\textwidth}{!}{%
\begin{tabular}{lc|cccccccc}
\toprule
                         & Train perct. & ERM    & Mixup & MLDG   & RSC   & AND-Mask & SimCLR  &  Fish  & Ours  \\\midrule
\multirow{6}{*}{\rotatebox{90}{DSADS}}   & 20\%      & 61.38 & 75.15 & 64.86 & 62.18 & 63.03    & 75.96 &  63.58 & \textbf{87.88} \\
                         & 40\%      & 64.46 & 82.48 & 67.25 & 67.70 & 65.59    & 75.76  &  65.82 & \textbf{89.71} \\
                             & 60\%      & 63.90 & 82.70 & 71.46 & 69.98 & 69.51    & 75.61  &  67.65 & \textbf{90.43} \\
                         & 80\%      & 67.19 & 81.58 & 73.34 & 75.37 & 69.38    & 74.69  &  66.03 & \textbf{90.97} \\
                         & 100\%     & 67.22 & 83.44 & 74.56 & 75.58 & 74.20    & 76.22  &  69.35 & \textbf{91.95} \\
                         & Avg       & 64.83 & 81.07 & 70.30 & 70.16 & 68.34    & 75.65 &  66.49 & \textbf{90.19} \\\midrule
\multirow{6}{*}{\rotatebox{90}{PAMAP2}}  & 20\%      & 67.45 & 71.74 & 71.58 & 71.45 & 70.41    & 73.57 &  72.85 & \textbf{82.95} \\
                         & 40\%      & 70.28 & 76.69 & 74.06 & 73.73 & 72.82    & 74.25 &  77.02 & \textbf{84.34} \\
                         & 60\%      & 72.23 & 77.83 & 75.26 & 75.72 & 74.28    & 74.71 &  76.04 & \textbf{85.03} \\
                         & 80\%      & 72.56 & 78.00 & 77.34 & 76.17 & 75.03    & 74.09 &  75.13 & \textbf{86.67} \\
                         & 100\%     & 74.09 & 79.72 & 78.97 & 77.96 & 75.10    & 74.25 &  75.49 & \textbf{86.31} \\
                         & Avg       & 71.32 & 76.80 & 75.44 & 75.01 & 73.53    & 74.17  &  75.31 & \textbf{85.06} \\\midrule
\multirow{6}{*}{\rotatebox{90}{USC-HAD}} & 20\%      & 64.88 & 67.22 & 62.12 & 70.17 & 66.06    & 67.22  &  67.42 & \textbf{77.36} \\
                         & 40\%      & 71.11 & 75.30 & 68.49 & 77.31 & 71.87    & 69.16 &  73.54 & \textbf{80.72} \\
                         & 60\%      & 74.67 & 78.14 & 71.60 & 77.59 & 74.31    & 71.38 &  76.09 & \textbf{80.88} \\
                         & 80\%      & 75.43 & 79.76 & 71.79 & 78.65 & 76.60    & 71.99  & 77.21 & \textbf{82.49} \\
                         & 100\%     & 74.90 & 81.27 & 72.05 & 79.41 & 76.01    & 72.14  &  78.92 & \textbf{82.51} \\
                         & Avg       & 72.20 & 76.34 & 69.21 & 76.62 & 72.97    & 70.38  &  74.64 & \textbf{80.80}\\\bottomrule
\end{tabular}
}
\vspace{-.2in}
\end{table}

\subsection{Robustness in Low-resource Scenarios~(RQ2)}
\label{exp-robustness}

\begin{figure}[htbp]
	\centering
	\subfigure[Improvement over ERM]{
		\includegraphics[width=0.47\linewidth]{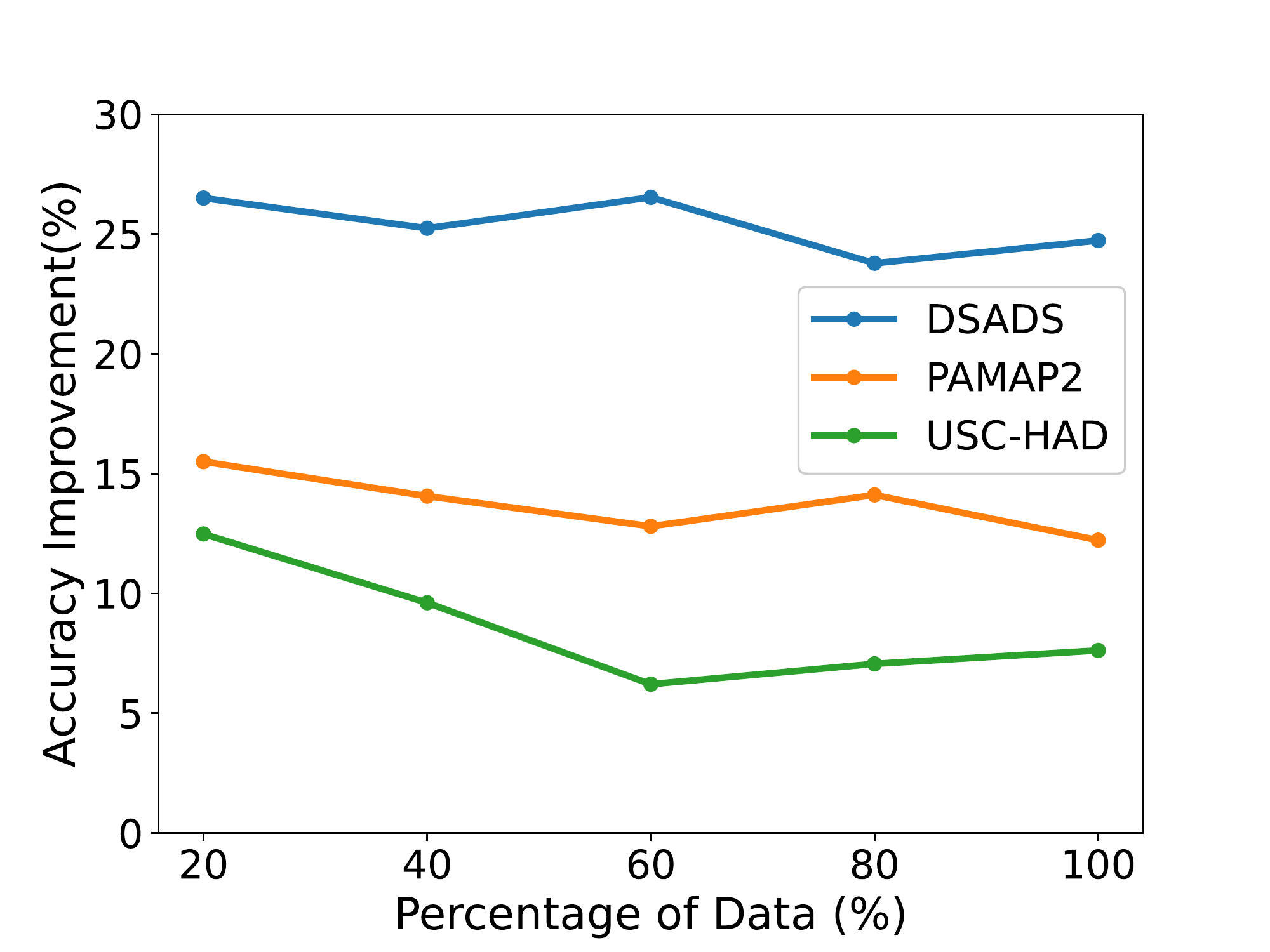}
		\label{fig-compare-erm}}
	\subfigure[Impro. over the second-best baseline]{
		\includegraphics[width=0.47\linewidth]{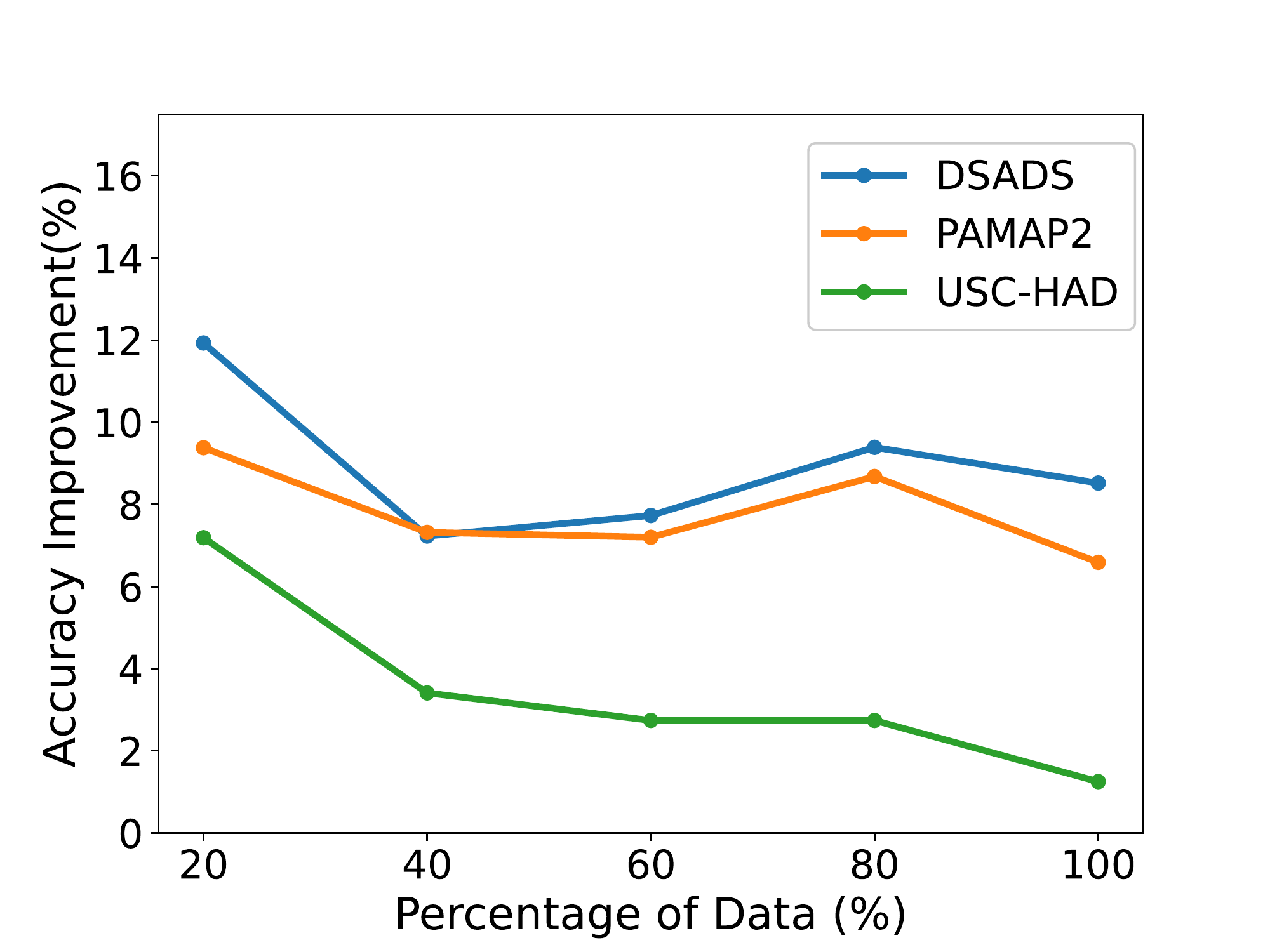}
		\label{fig-compare-othermethod}}
	\vspace{-.1in}
	\caption{Results improvement over ERM (baseline) and the second-best result with different percentage of training data.}
	\vspace{-.2in}
	\label{fig-different-data-improvment}
	
\end{figure}

To evaluate the robustness of the proposed method, we make a comparison under different amounts of training data by varying the training data proportions: i.e., $\{100\%, 80\%, 60\%, 40\%, 20\%\}$ of training data.
The results are shown in \tablename~\ref{tb-differ-data-rate}.
Results demonstrate that with different percentages of training data, the proposed \method can attain the best results with significant accuracy improvement over other baselines.
As the amount of data continues to decrease, the accuracy of all methods tends to decline.
Specifically, our \method suffers performance drops of 4.07\%, 3.36\%, and 5.15\% on three datasets when the training data proportion drops from 100\% to 20\%, which the drops are 8.29\%, 7.98\%, and 14.05\% for the second-best baselines.
This indicates the superiority of our approach.
Besides, \figurename~\ref{fig-different-data-improvment} implies that the improvement is more obvious both compared with the baseline ERM and the second-best baselines with data reduction.
Especially, the improvement can reach 26.5\% and 11.93\% compared with ERM and the best comparison method on DSADS.
This is because DSADS has the smallest amount of samples among the three datasets.
These observations illustrate that our method is superior and more robust when faced with low-resource generalizable HAR problems. 

\subsection{Ablation Study and Interpretation~(RQ3)}
\label{exp-ablation}

We conduct quantitative and qualitative analysis to investigate the contribution of each module in \method and more importantly, to interpret why \method is effective for low-resource generalizable HAR problems.
We compare the complete version \method with five variants, i.e., ERM; ERM+Augmentation~(ERM+A) which uses the original and augmentation data to train a deep model with ERM; ERM+Diversity Generation~(ERM+DG); ERM+Diversity Generation+Diversity Preservation~(ERM+DG+DP); and ERM+Diversity Generalization+Discrimination Enhancement~(ERM+DG+DE).

\begin{figure}[t]
    \centering
\vspace{-.2in}
    \includegraphics[width=0.65\linewidth]{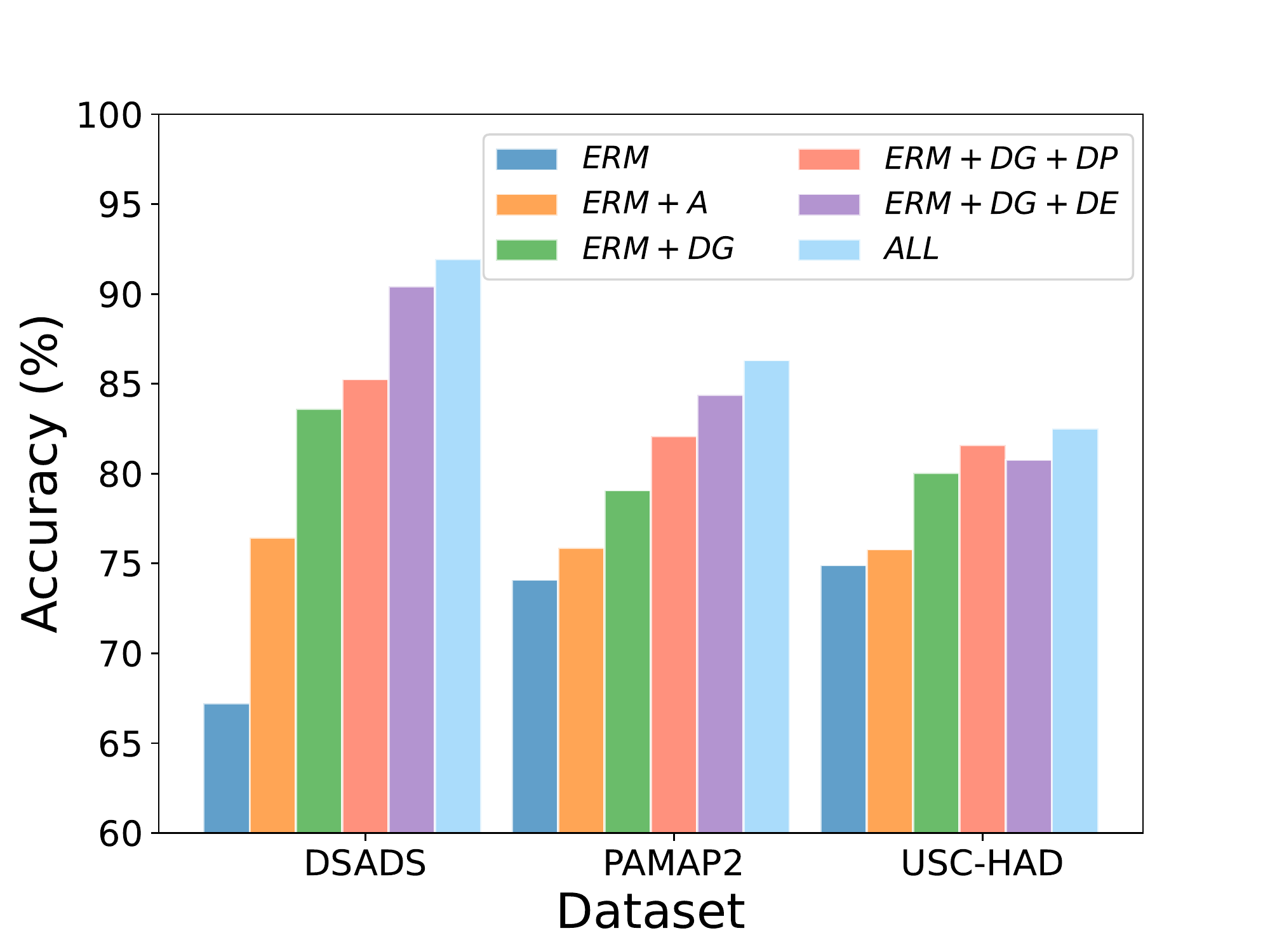}
 \vspace{-.1in}
    \caption{Ablation study of \method.}
 \vspace{-.2in}
\label{fig-ablation}
\end{figure}

\paragraph{Quantitative ablation results.}
Results are shown in Figure~\ref{fig-ablation}.
Comparing ERM and ERM+A, we see that data augmentation can help improve the activity recognition accuracy, especially on the DSADS dataset where there is a significant improvement of 9.21\%.
This is because DSADS has the smallest number of samples among the three datasets, indicating that the augmentation is more effective for the low-resource scenario.
With the diversity generation module, the performance of the model is highly improved by 3.21\% to 7.17\%.
This indicates the self-supervised auxiliary task can help a lot in exploring latent diversified properties.
By adding diversity preservation and discrimination enhancement, the classification accuracy is further improved, which illustrates that both modules make contributions to the HAR task.
Combining all modules, the proposed \method can achieve the best performance for the low-resource generalizable HAR tasks.

\begin{figure}[t!]
	\centering
	\subfigure[ERM]{
		\includegraphics[width=0.14\textwidth]
		{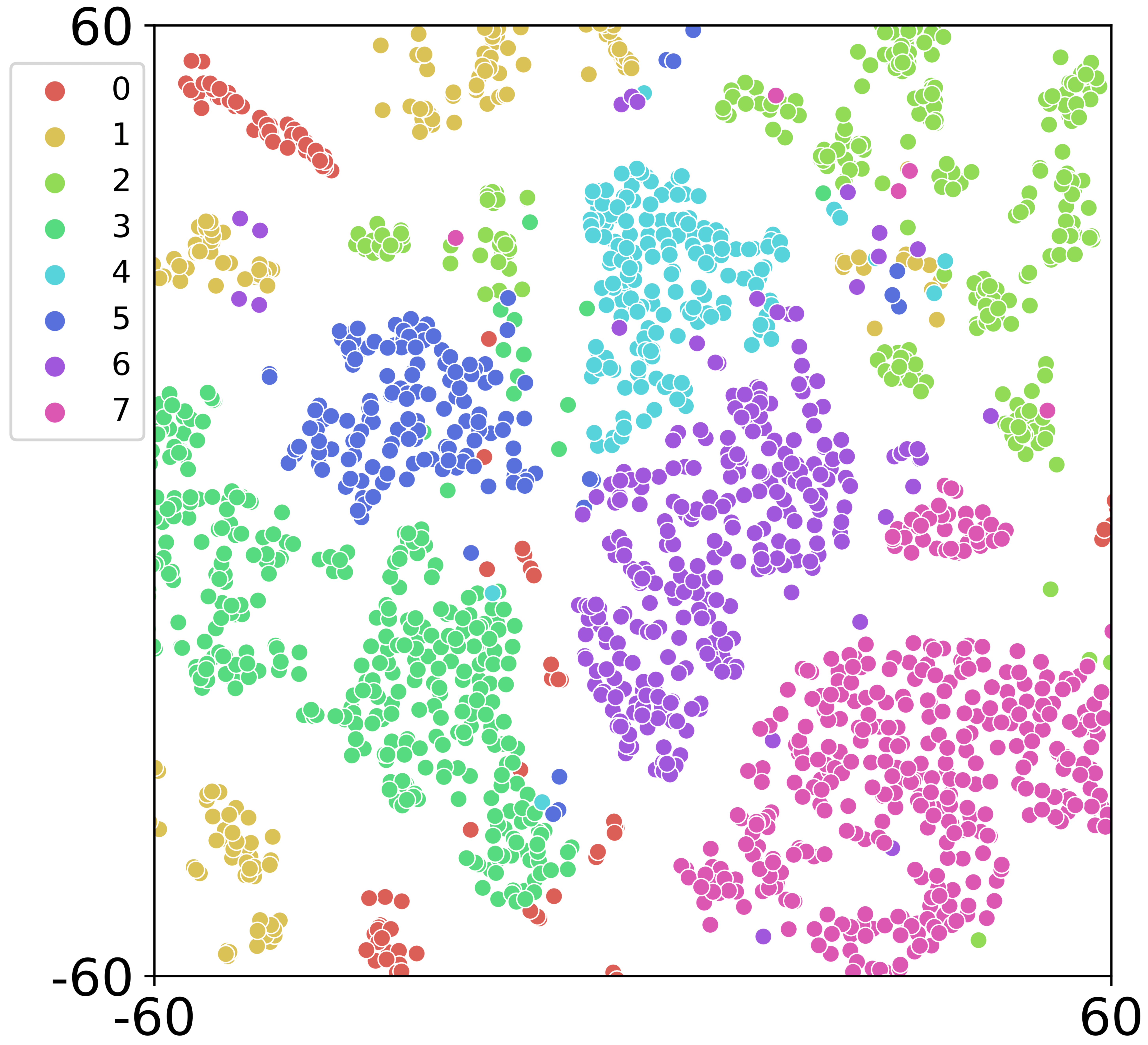}
		\label{fig-tsne-erm}}
	\subfigure[ERM+A]{
		\includegraphics[width=0.14\textwidth]
		{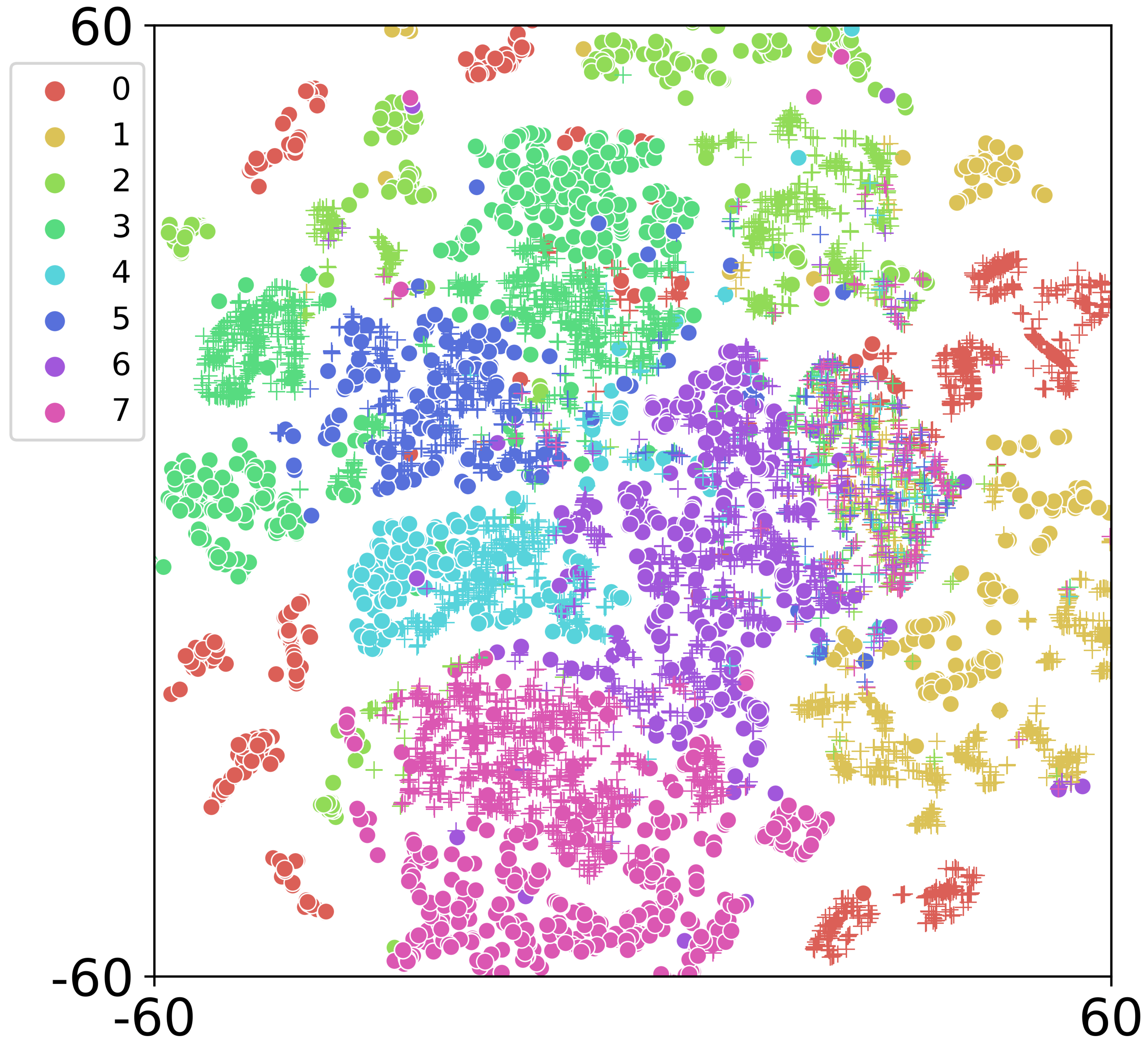}
		\label{fig-tsne-erm-a}}
	\subfigure[ERM+DG]{
		\includegraphics[width=0.14\textwidth]
		{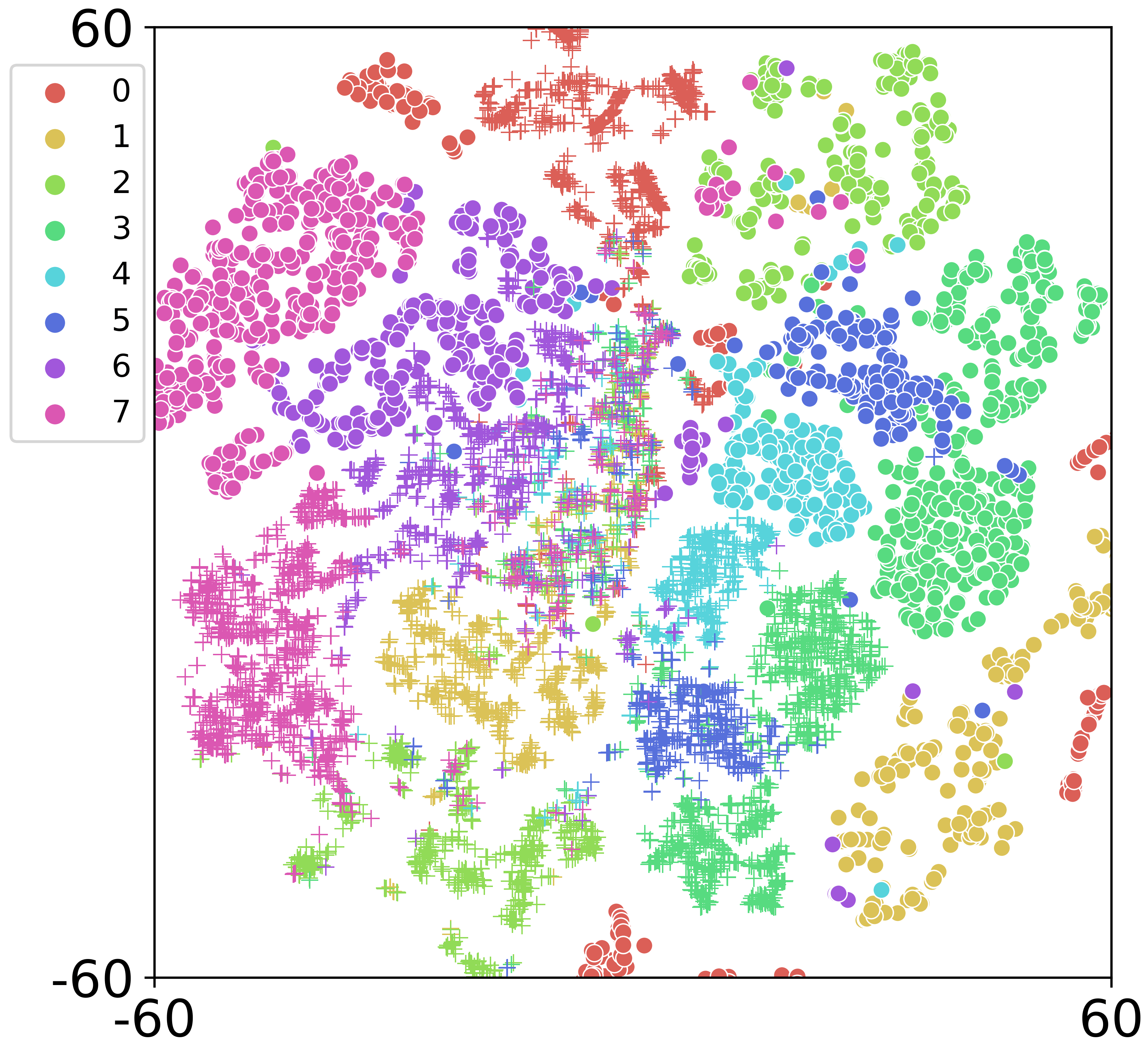}
		\label{fig-tsne-erm-dg}}
	\subfigure[ERM+DG+DP]{
		\includegraphics[width=0.14\textwidth]
		{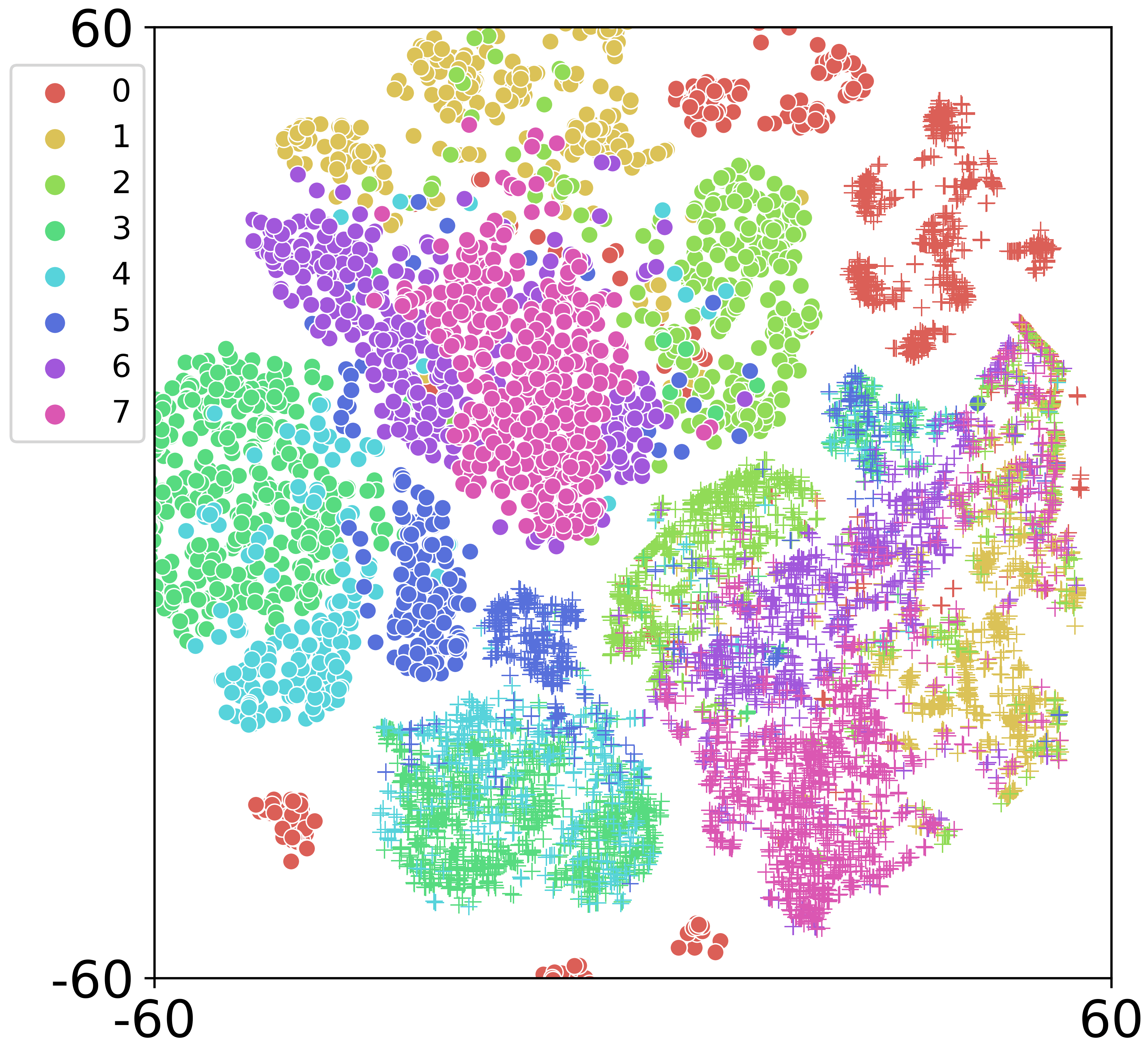}
		\label{fig-tsne-wode}}
	\subfigure[ERM+DG+DE]{
		\includegraphics[width=0.14\textwidth]
		{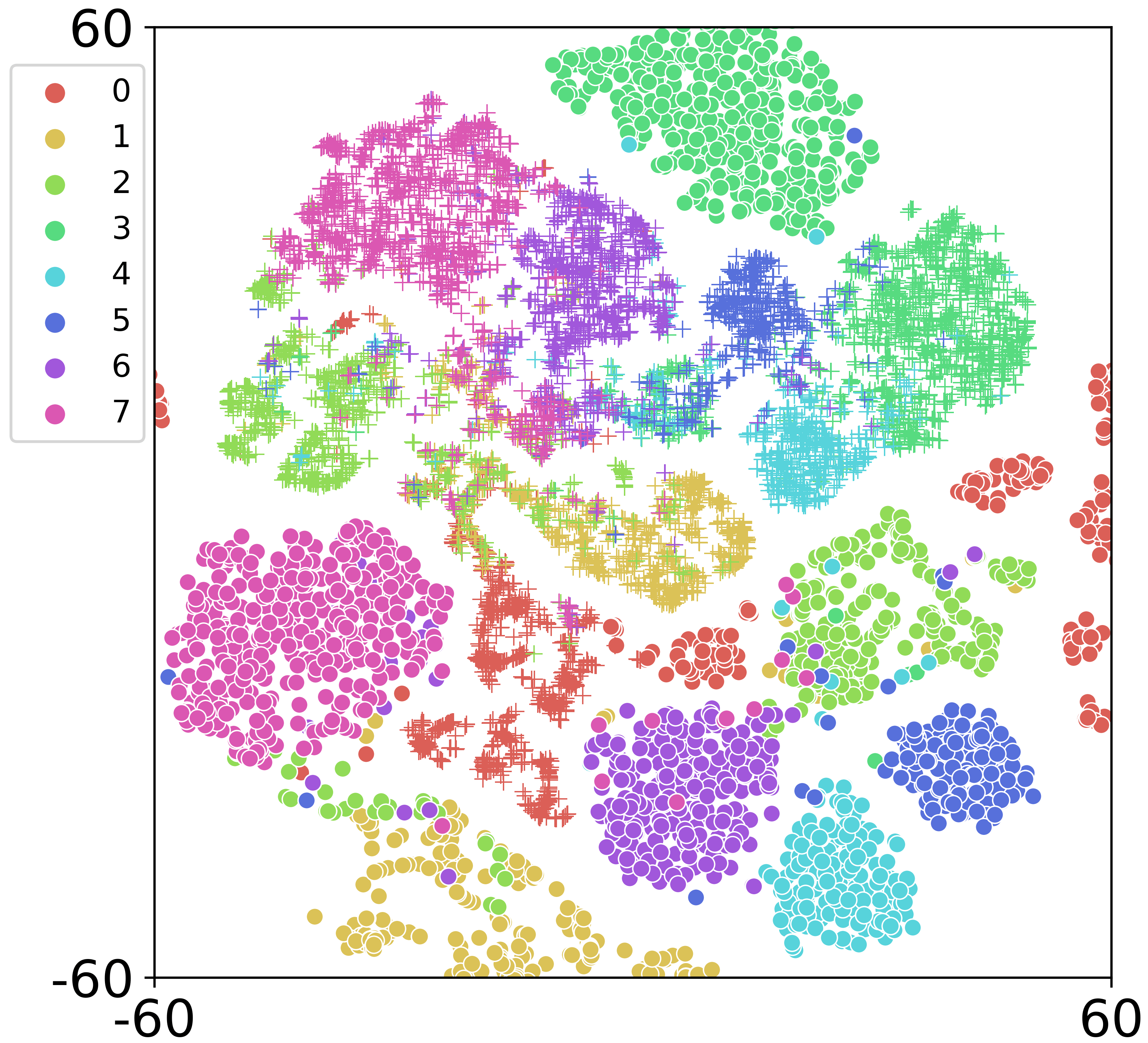}
		\label{fig-tsne-wodp}}
	\subfigure[\method]{
		\includegraphics[width=0.14\textwidth]
		{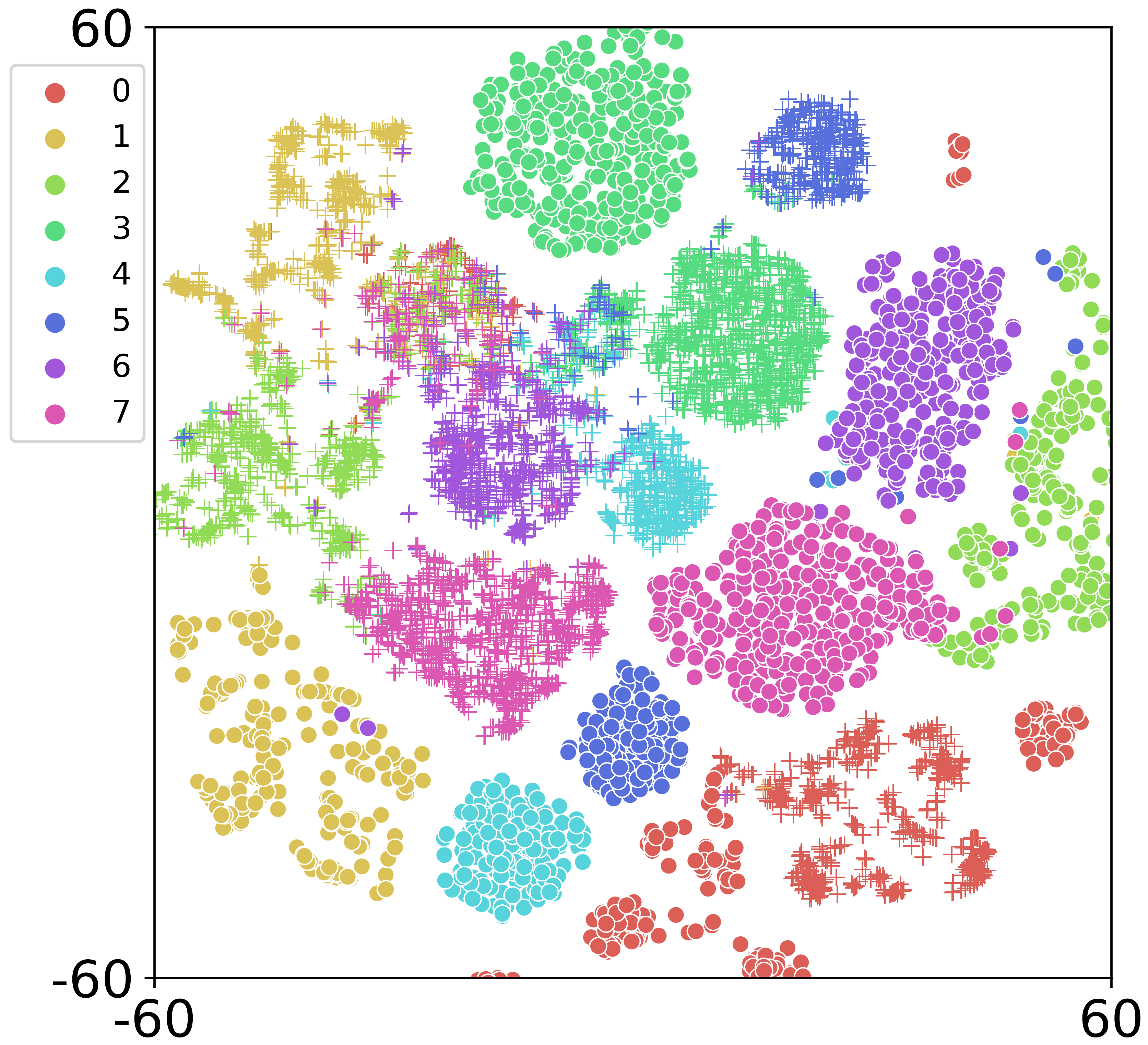}
		\label{fig-tsne-all}}
	\vspace{-.1in}
	\caption{Visualization of the t-SNE embeddings of the PAMAP2 dataset. We randomly select the same amounts of original and augmented data. Each class is denoted by color. The original and augmented domains are denoted by shapes of dot and plus. The classes are denoted by numbers: lying, sitting, standing, walking, ascend stairs, descend stairs, vacuum cleaning, and iron. \textsl{Best viewed in color and zoom in.}}
	\label{fig-tsne}
	\vspace{-.2in}
\end{figure}

\begin{figure*}[t!]
	\centering
	\vspace{-.1in}
	\subfigure[ERM]{
		\includegraphics[width=0.28\textwidth]{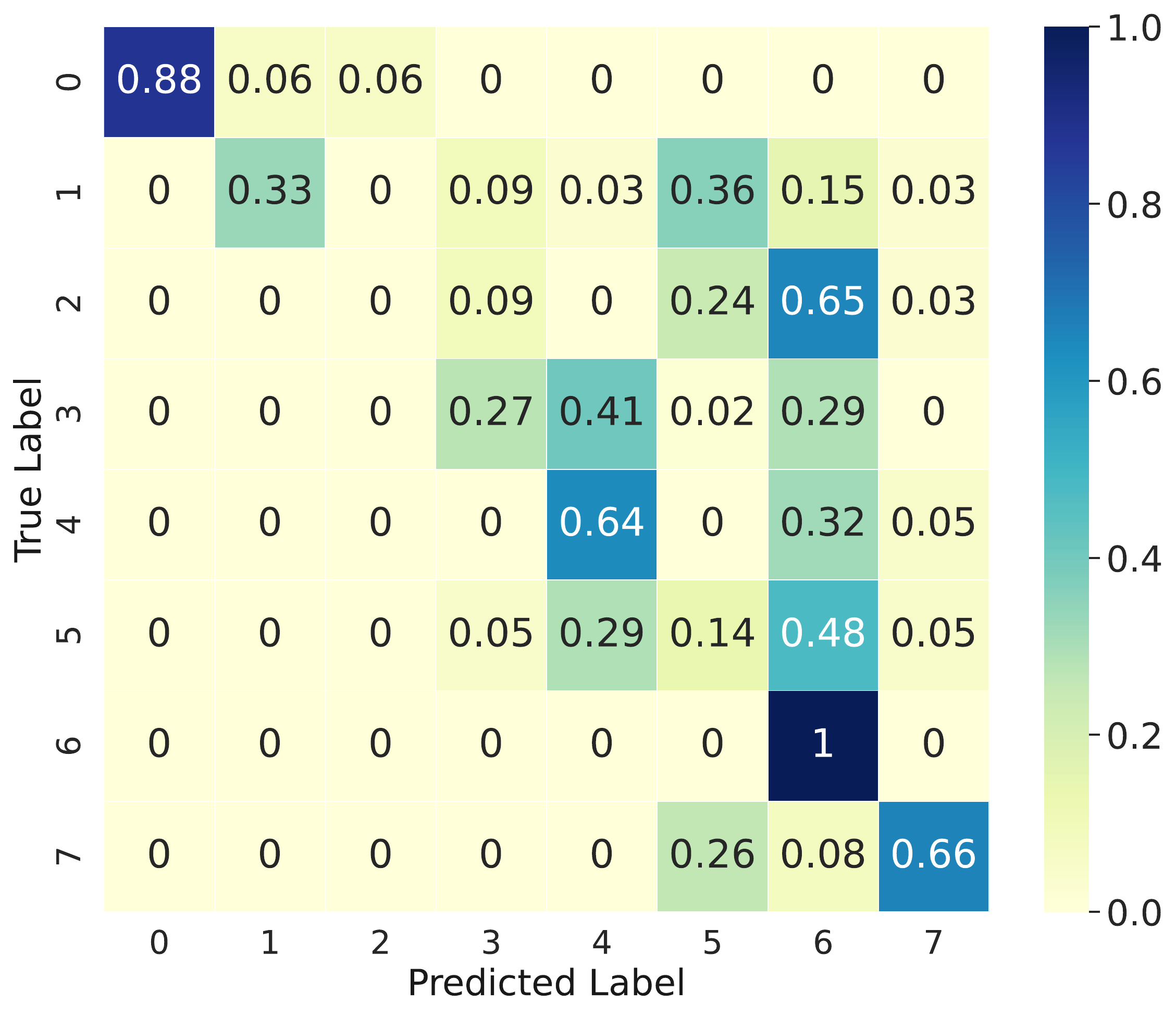}
		\label{fig-conf-erm}}
		\hspace{.2in}
	\subfigure[Mixup]{
		\includegraphics[width=0.28\textwidth]{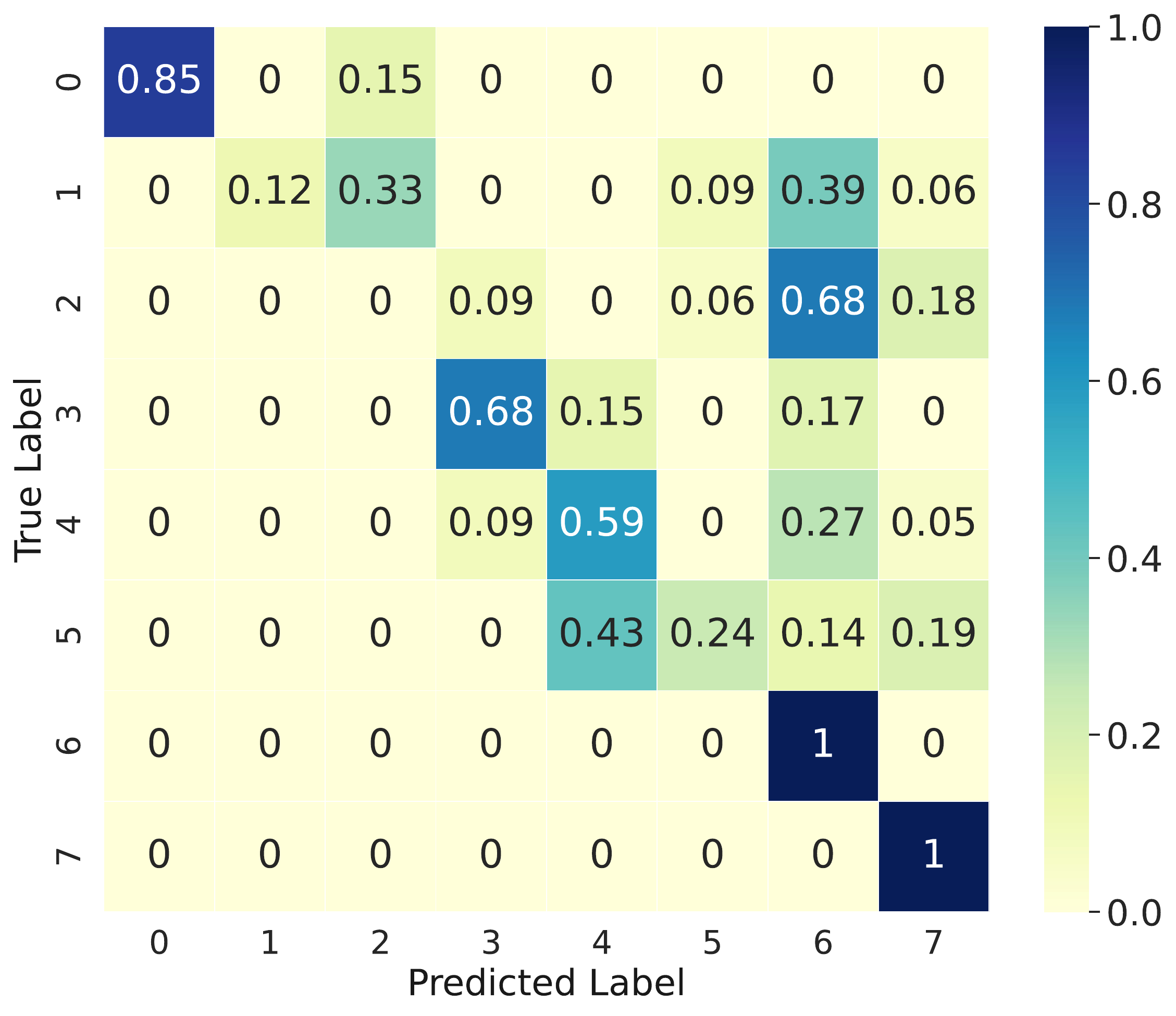}
		\label{fig-conf-mixup}}
		\hspace{.2in}
	\subfigure[\method (ours)]{
		\includegraphics[width=0.28\textwidth]{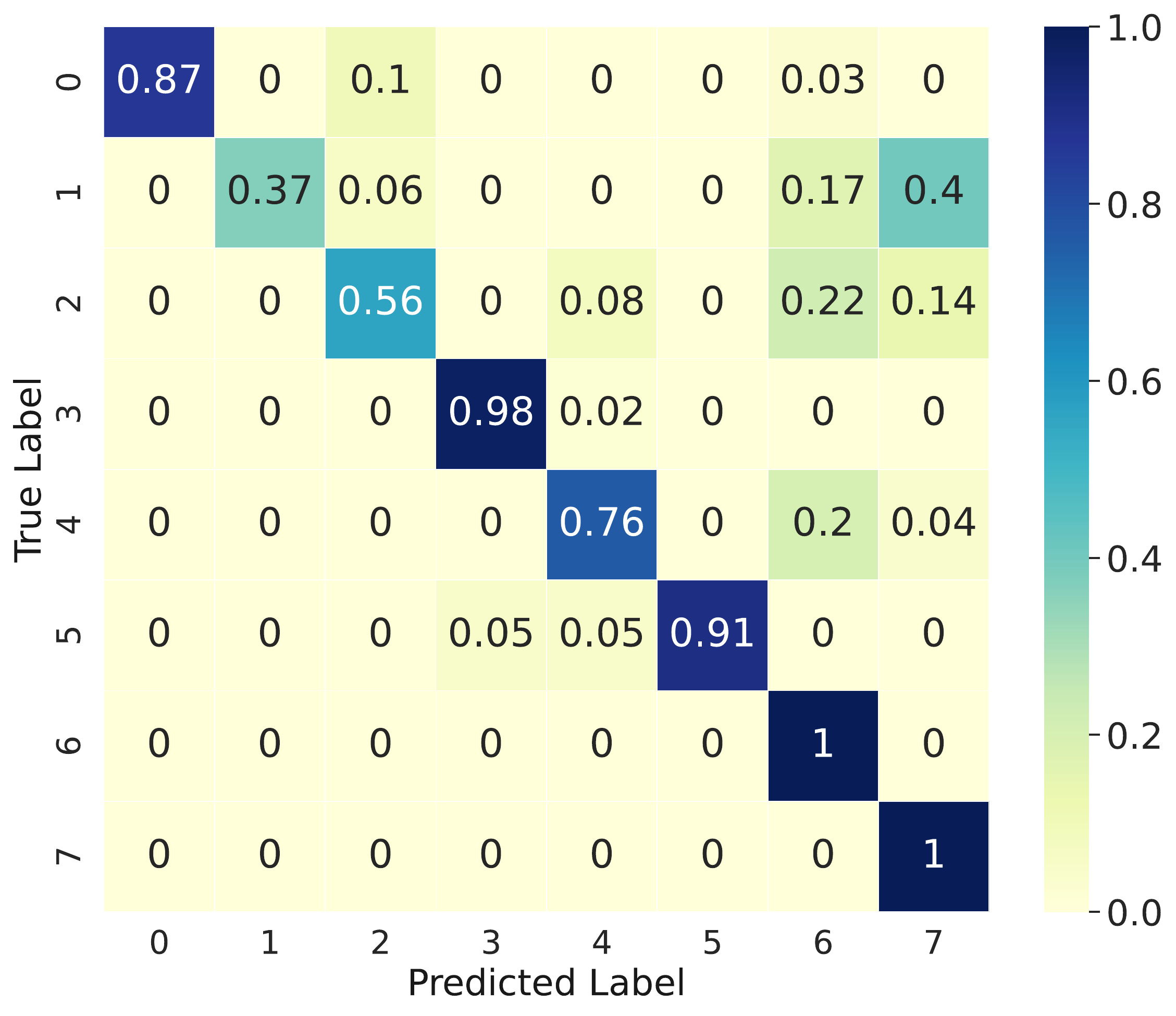}
		\label{fig-conf-my}}
	\vspace{-.2in}
	\caption{The confusion matrices of the first task on PAMAP2 dataset. Labels 0-7 denote activities: lying, sitting, standing, walking, ascend stairs, descend stairs, vacuum cleaning, and iron.}
	\label{fig-confusion}
\end{figure*}

\begin{figure*}[t!]
  \centering
   	\subfigure[F1-score]{
		\includegraphics[width=0.22\linewidth]{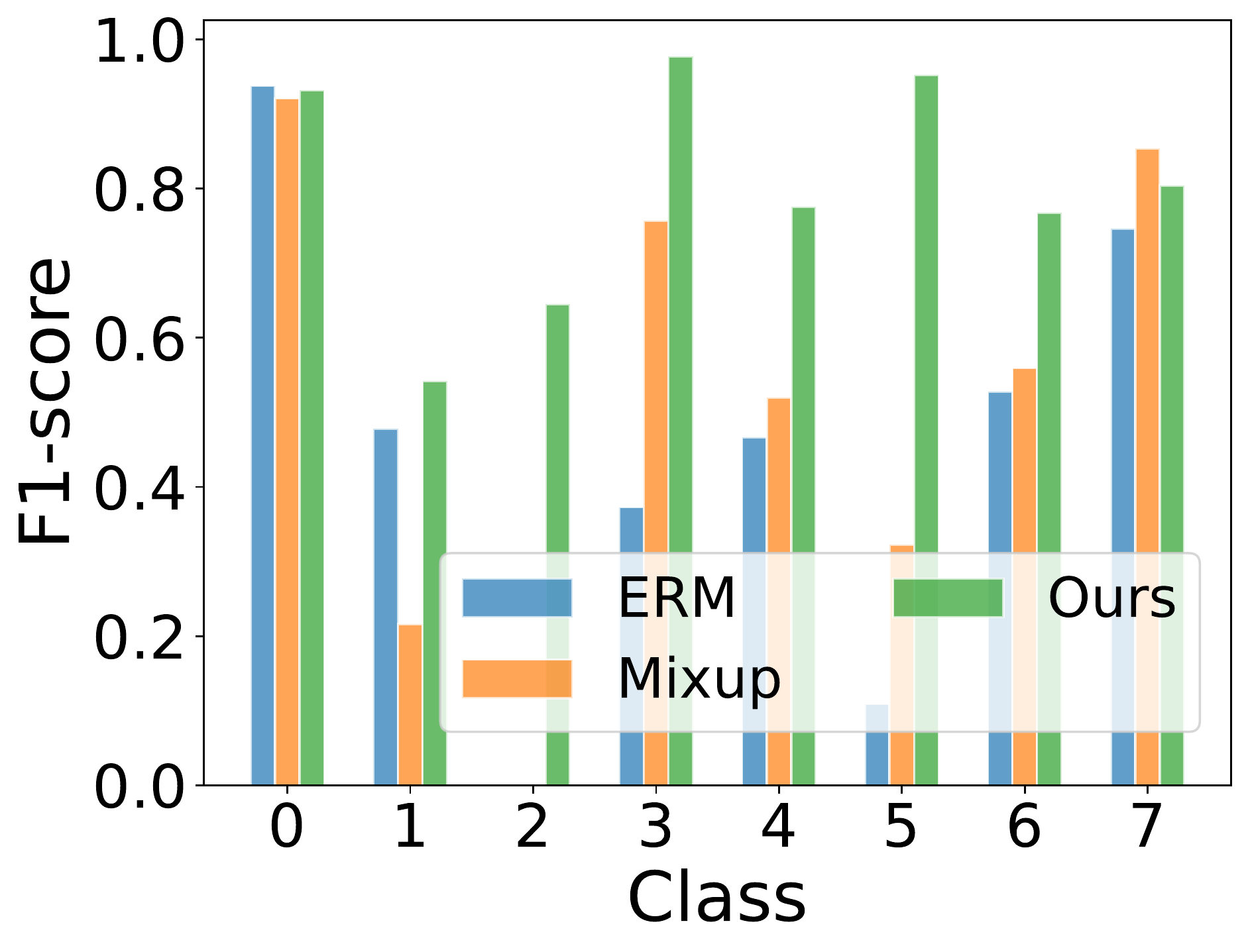}
		\label{fig-f1}}
  \subfigure[Precision]{
		\includegraphics[width=0.22\linewidth]{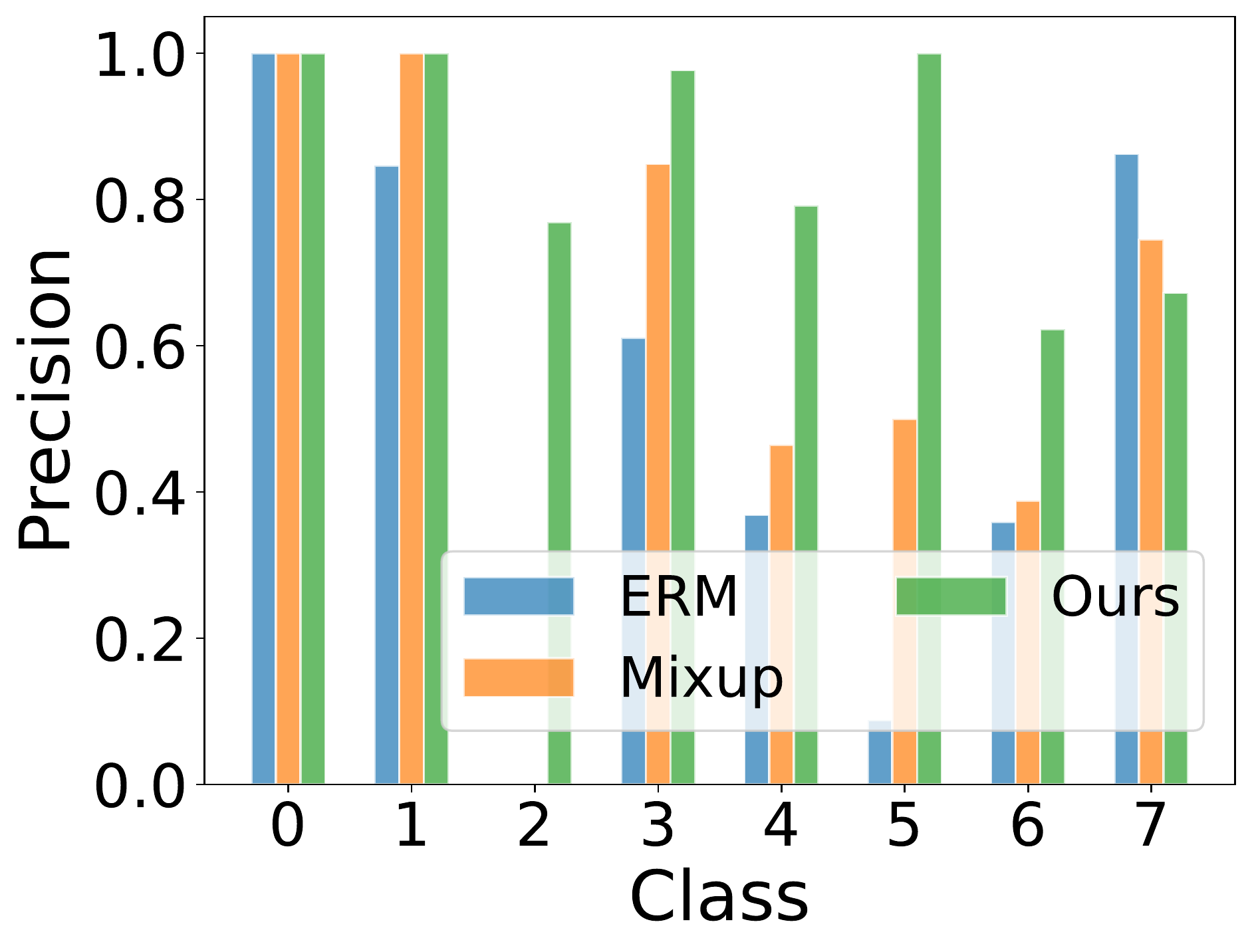}
		\label{fig-pre}}
	\subfigure[Recall]{
		\includegraphics[width=0.22\linewidth]{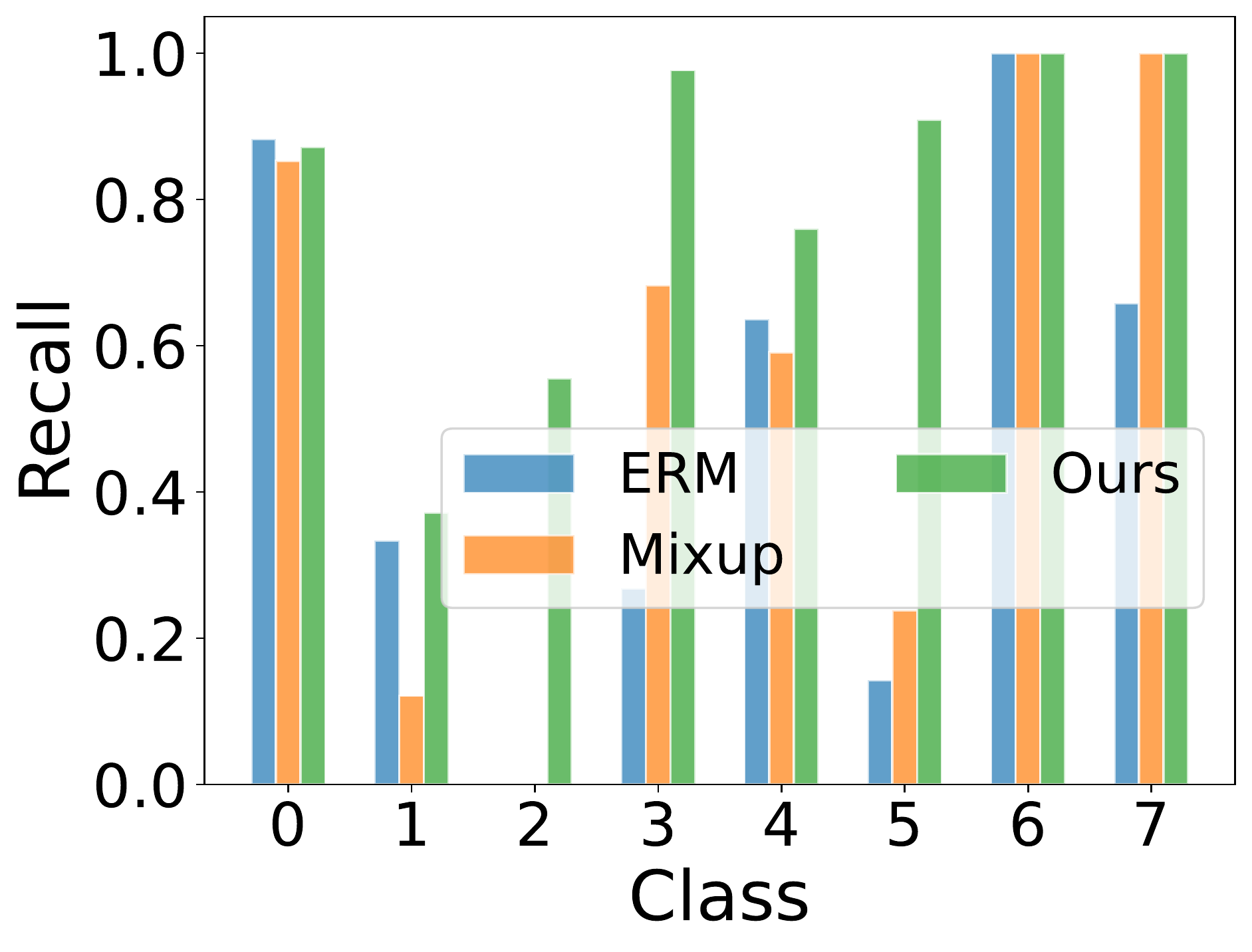}
		\label{fig-recall}}
\rulesep
    \subfigure[Compatibility]{
        \includegraphics[width=0.22\linewidth]{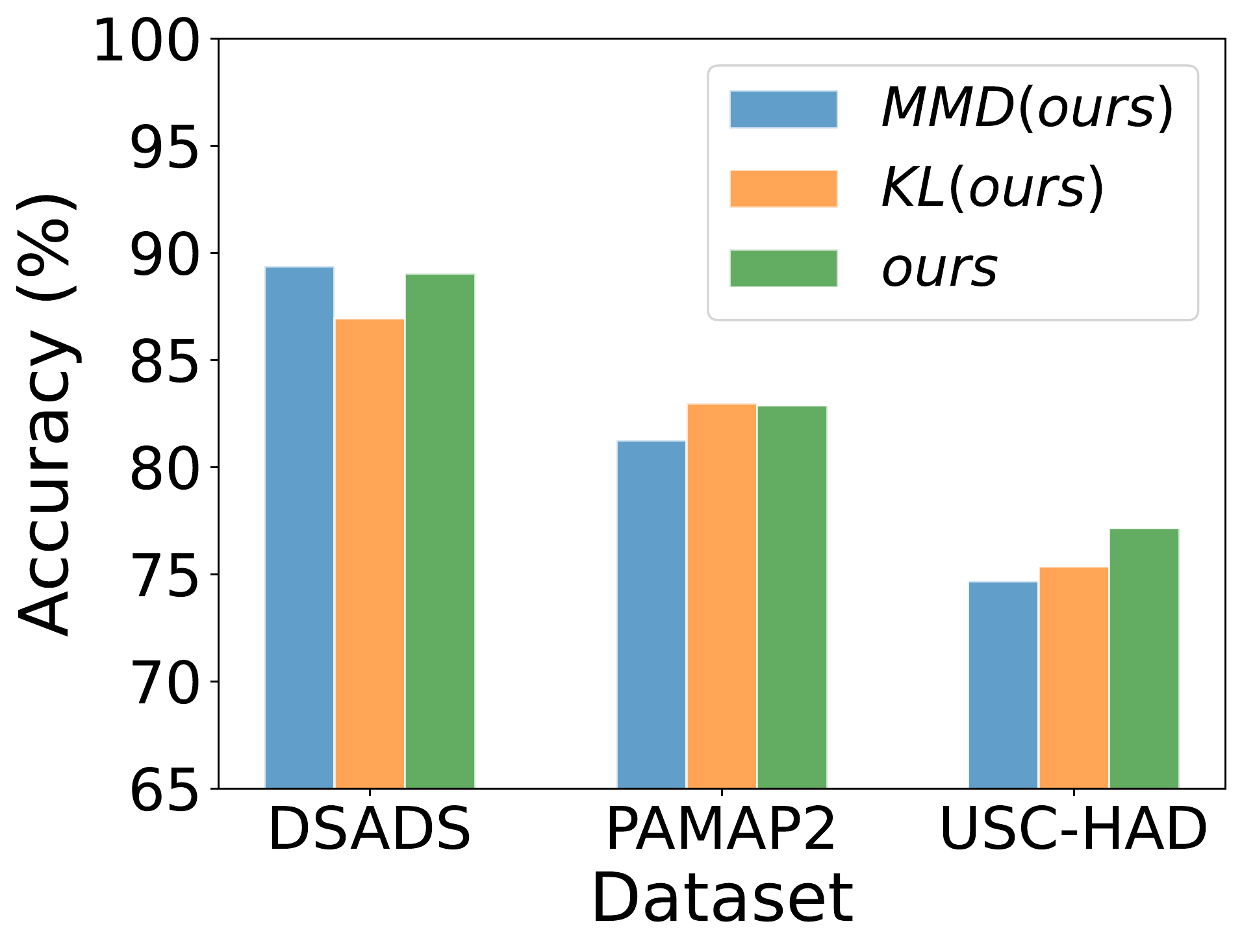}
        \label{fig-extend-dist}
  }
  \vspace{-.2in}
    \caption{(a)(b)(c) Class-wise F1-score, precision, and recall. (d) Compatibility.}
  \label{fig-f1-extend}
 \vspace{-.1in}
\end{figure*}

\paragraph{Qualitative ablation results.}

For more interpretable analysis, we further visualize the feature embeddings to show the effect of each component in \method using t-distributed stochastic neighbor embedding~(t-SNE)~\cite{van2008visualizing} on the PAMAP2 dataset with $100\%$ data.
From \figurename~\ref{fig-tsne}, we have the following observations:
(1) Comparing ERM+A with ERM, the data space is largely expanded by using data augmentation and the gaps between intra- and inter-class samples are filled up with diversified augmented data.
This indicates that using augmentation can expand the representation space with more diversity.
(2) As shown in \figurename~\ref{fig-tsne-erm-dg}, when adding the self-supervised auxiliary task, the class discrimination gets enhanced.
This might be due to the exploration of some latent characteristics of activities.
(3) \figurename~\ref{fig-tsne-wode} shows that by introducing the domain discrimination between the original and augmented data, these two domains are pulled away that avoids them to overlap, which preserves the diversity.
(4) From \figurename~\ref{fig-tsne-wodp}, by using contrastive learning, the margins between different classes are becoming larger and representations are more discriminated.
(5) Combining all components, \method can attain better representations that are more diverse and more semantic-discriminated to achieve accurate activity recognition performance.

\subsection{Case Study by Class-Wise Analysis~(RQ4)}
\label{exp-casestudy}

We further evaluate the classification performance of each activity as case study.
We utilize the confusion matrix and class-wise precision, recall, and F1-score to make a fine-grained analysis.
We compare our approach with the baseline method ERM~\cite{vapnik1992principles} and a state-of-the-art method Mixup~\cite{xu2020adversarial} on the first task of PAMAP2 dataset with 100\% training data.

The confusion matrices are shown in \figurename~\ref{fig-confusion} and \figurename~\ref{fig-f1}-\ref{fig-recall} present the F1-score, precision, and recall of each class.
From the confusion matrices, we can observe that directly minimizing the classification error on the training data with ERM may get degraded performance on the unseen test data.
Mixup has better generalization performance with data augmentation while has less satisfactory performance on certain activities such as sitting and standing.
This may be because it only enlarges the data diversity while lacking semantic discrimination capability.
Our \method can improve the poorly-performed classes by considering both diversity and discrimination learning.
Thus as shown in \figurename~\ref{fig-f1}-\ref{fig-recall}, our approach gets the best F1-score, precision and recall on most activities.
Specifically, it significantly improves the performance on difficult category \textit{standing} that is misclassified by other approaches.
This case study demonstrates the effectiveness of \method.

\subsection{Compatibility~(RQ5)}
\label{sec-exp-exten}

What about the compatibility of \method?
As mentioned in Section~\ref{sec-dp}, \method is a general framework and can be feasibly extended with other distance metrics for diversity preservation.
In this section, we replace the domain discriminator with two typical distance metrics, namely Maximum Mean Discrepancy~(MMD)~\cite{gretton2012kernel} and Kullback-Leibler~(KL) divergence~\cite{kullback1997information}.
MMD calculates the discrepancy between embeddings in Reproducing Kernel Hibert Space~(RKHS) and can be regarded as a technique to measure the distribution distance between two domains.
KL divergence is also a common technique to measure domain similarity.

As shown in \figurename~\ref{fig-extend-dist}, \method obtains similar results by replacing minimizing loss of domain discriminator with maximizing MMD and KL, which are clearly better than previous comparison methods.
This indicates that \method can be feasibly extended with different distance metrics and it achieves the best average accuracy on three datasets with domain discriminator. Thus, we mainly use this one in our method.

\begin{figure}[t!]
	\centering
	\subfigure[$\lambda$]{
		\includegraphics[width=0.32\linewidth]{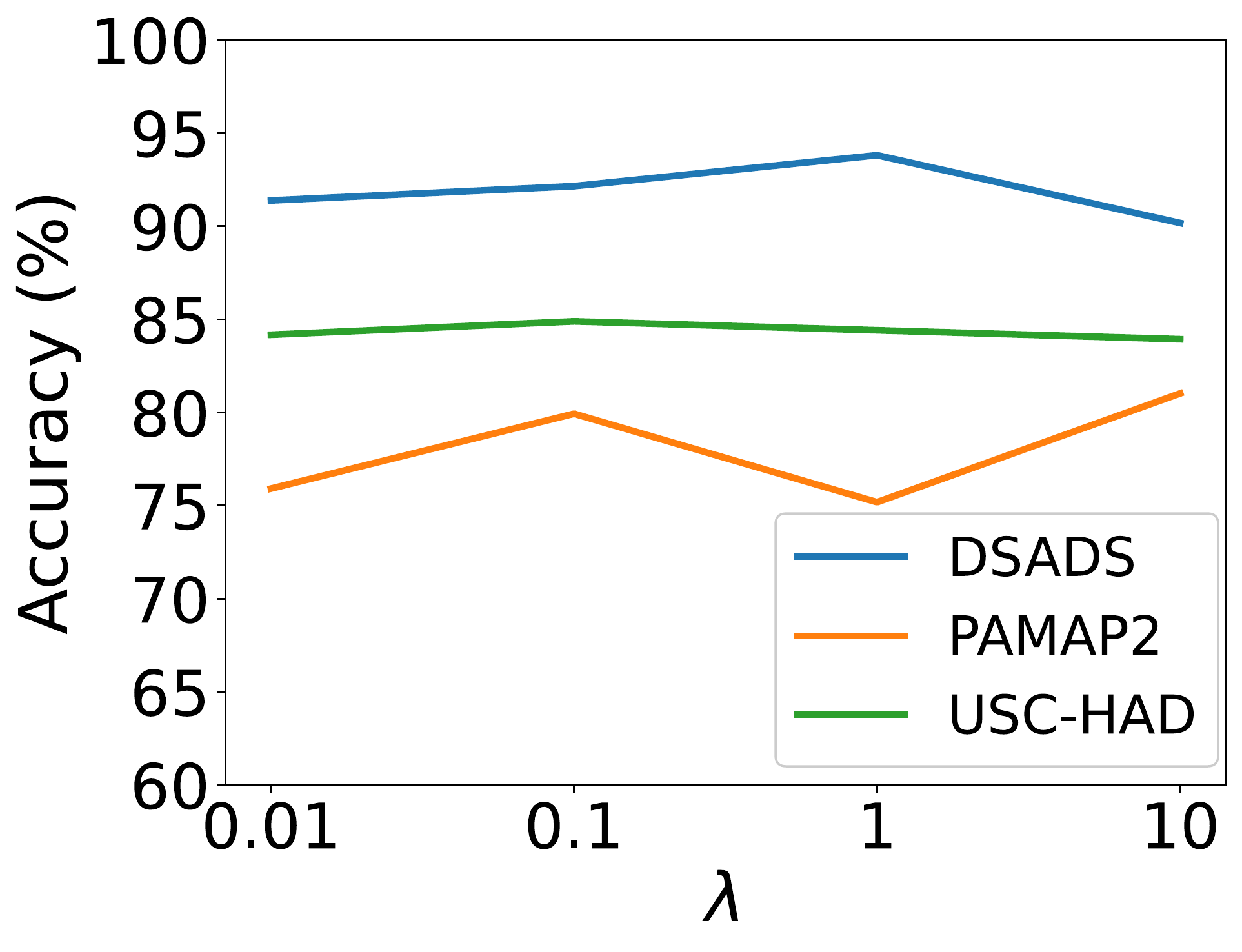}
		\label{fig-lambda}}\hspace{-.1in}
	\subfigure[$\beta$]{
		\includegraphics[width=0.32\linewidth]{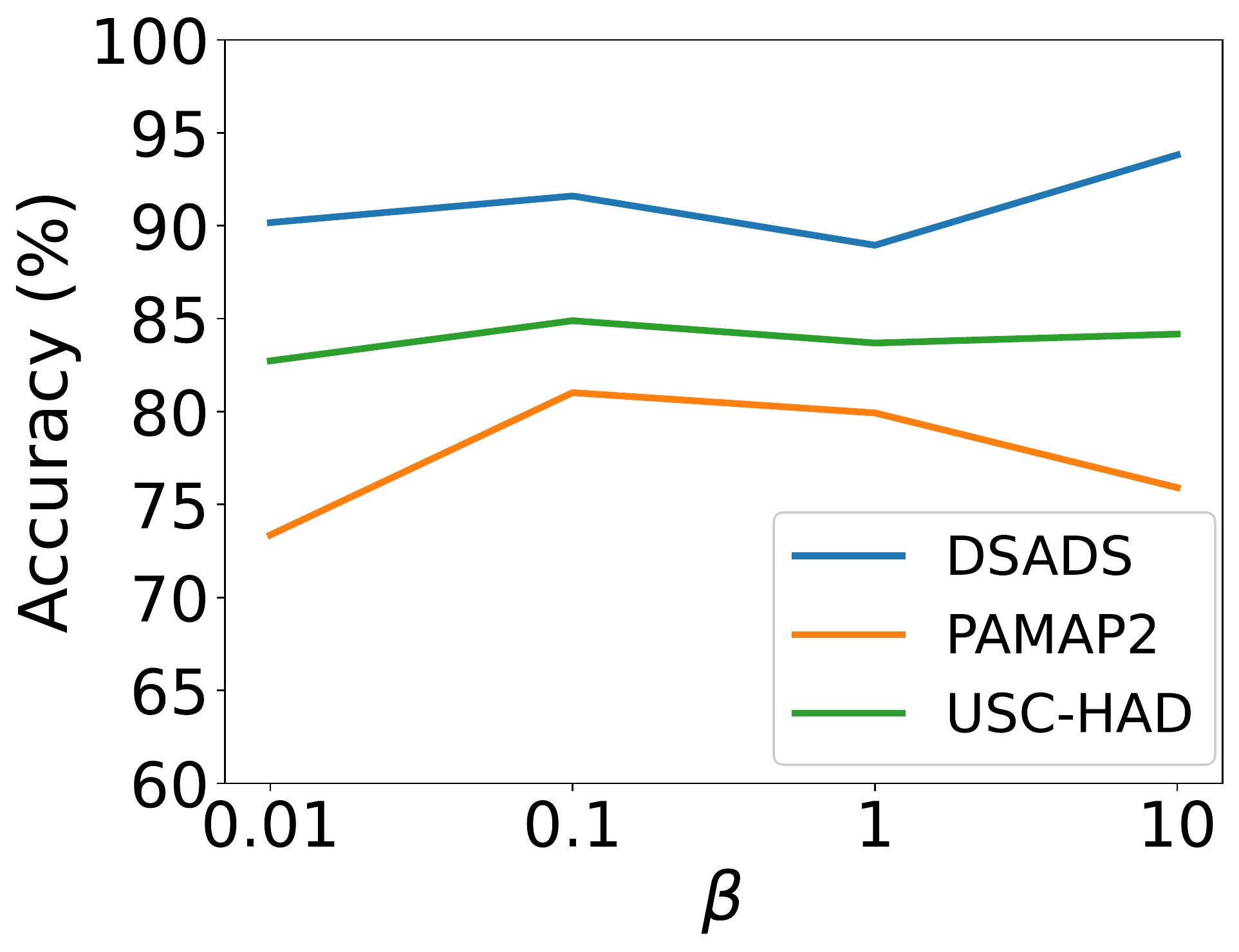}
		\label{fig-beta}}\hspace{-.1in}
	\subfigure[$\gamma$]{
		\includegraphics[width=0.32\linewidth]{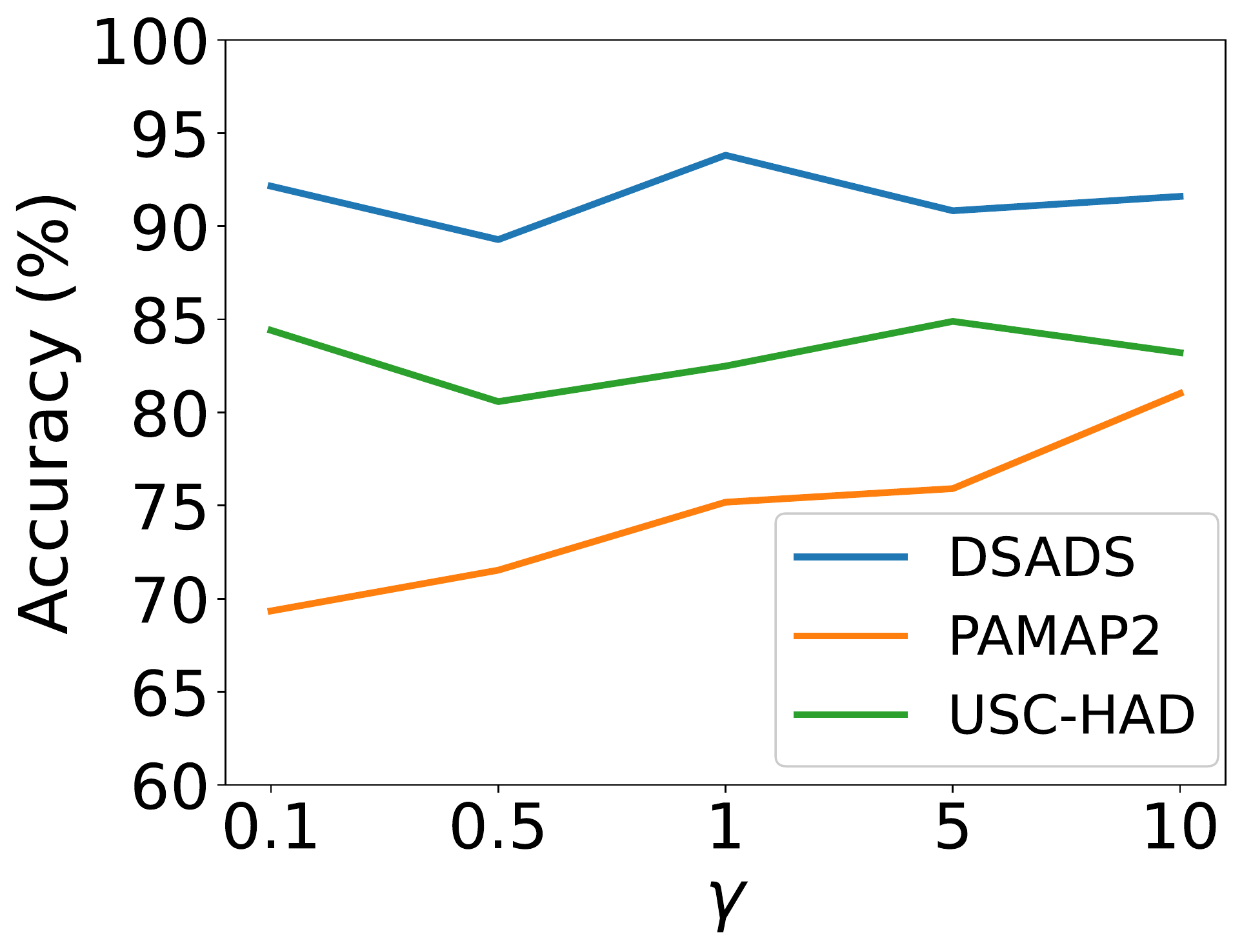}
		\label{fig-gamma}}
	\caption{Parameter sensitivity analysis}
	\label{fig-paramsen}
\end{figure}

Additionally, we provide parameter sensitivity analysis. We focus on three key trade-off hyper-parameters of each learning module: $\lambda$ for diversity generalization module, chosen value from $\lambda \in \{0.01, 0.1, 1, 10\}$, $\beta$ for diversity preservation module chosen from $\beta \in \{0.01, 0.1, 1, 10\}$ and $\gamma$ for discrimination enhancement module, chosen from $\gamma \in \{0.1, 0.5, 1, 5, 10\}$.
As shown in Figure~\ref{fig-paramsen}, \method~has robust performance with a wide range of hyperparameters on three public datasets.

\section{Conclusions and Future Work}
\label{sec-con}

In this paper, we proposed \method for low-resource generalizable HAR.
\method can generate diversity in data space and explore the latent activity properties.
Then, feature diversity is further preserved by enlarging the distribution divergence between the original and the augmented domains.
By utilizing supervised contrastive learning, it can enhance the semantic discrimination of features.
\method significantly outperformed SOTA methods in comprehensive experiments which is a generic, explainable, and flexible framework.

In the future, we will use \method to assist in mining deep correlation between the sensor-based data and motion-related diseases as well as optimization of wearable healthcare applications. 
Additionally, we will further equip \method with federated learning to take care of privacy issues for some safety-critical and privacy-related applications.

\section*{Acknowledgements}

This work is supported by the National Key Research and Development Plan of China No. 2021YFC2501202, Beijing Municipal Science \& Technology Commission (No. Z221100002722009), Natural Science Foundation of China (No. 61972383), the Strategic Priority Research Program of Chinese Academy of Sciences (No. XDA28040500).


\bibliographystyle{ACM-Reference-Format}
\bibliography{refs}


\appendix

\section{Details on Experimental Setup}

\subsection{Dataset Statistical Information and Pre-Processing}
\label{app-preprocess}
The statistical information of datasets are summarized in Table~\ref{tb-dataset}.
The sliding window is utilized to segment data, and we set the time duration of a window to 5 seconds~\cite{barshan2014recognizing} for DSADS, 5.12 seconds~\cite{reiss2012introducing} for PAMAP2, and 5 seconds for USC-HAD with an overlap rate of 50\% between consecutive windows, respectively. According to the sampling rate of each dataset: 25Hz for DSADS, 100Hz for PAMAP2, and 100Hz for USC-HAD, the window length is 125 readings, 512 readings, and 500 readings, respectively. We normalize data with MinMaxScaler and transform a 1D time series sample into a 2D shape with a height equals to 1. The shape of each batch is $(b, c, h, w)$, where $b$ is the cardinality of a mini-batch, $c$ is the number of channels which is corresponding to the total axes of sensors, $h$ is the height, and $w$ is window length. More detailed pre-processing settings are summarized in Table~\ref{tb-dataset-process}, which is helpful for reproducibility.

\begin{table*}[htbp]
  \caption{Statistical information summary of three public activity recognition datasets}
  \vspace{-.1in}
  \label{tb-dataset}
  \resizebox{.8\textwidth}{!}{
\begin{tabular}{ccccccc}
\toprule
Dataset & \#Subject & \#Sample    & \#Sampling Rate                                                           & Position                                                                            & Sensor                                                                                 & \#Activity \\ \midrule
DSADS   & 8         & $\sim$1.14M & 25Hz                                                                      & \begin{tabular}[c]{@{}c@{}}5 (torso, right arm,\\  left arm,\\  right leg, left leg)\end{tabular} & \begin{tabular}[c]{@{}c@{}}5 IMUs\\ (acc, gyro, mag)\end{tabular}                         & 19         \\ \hline
PAMAP2  & 9         & $\sim$2.84M & \begin{tabular}[c]{@{}c@{}}100Hz \end{tabular} & 3 (hand, chest, ankle)                                                                            & \begin{tabular}[c]{@{}c@{}}3 IMUs\\ ($\pm6g,\pm16g$ acc, gyro, mag),\\ a HRM\end{tabular} & 18         \\\hline 
USC-HAD & 14        & $\sim$2.81M & 100Hz                                                                     & 1 (front right hip)                                                                               & \begin{tabular}[c]{@{}c@{}}1 Motion Node\\ (acc, gyro)\end{tabular}                       & 12         \\ \bottomrule
\end{tabular}}
\end{table*}

\begin{table*}[htbp]
  \caption{Summary of data pre-process settings}
  \vspace{-.1in}
  \label{tb-dataset-process}
  \resizebox{\textwidth}{!}{
\begin{tabular}{cccccccccc}
\toprule
Dataset & \#Task & \#Subject                                                          & \#Activity & Sensors                                                        & Win. size & Win. duration & Sample shape & Kernel size & Network                                                                                         \\ \midrule
DSADS   & 4        & 8                                                                       & 19              & \begin{tabular}[c]{@{}c@{}}acc,\\ gyro,\\  mag\end{tabular}    & 125           & 5s              & (45, 1, 125) & (1, 9)      & \begin{tabular}[c]{@{}c@{}}2 conv2d layers\\ (Relu-MaxPool2d)\\ 1 fc layer\\ (Relu)\end{tabular} \\\hline
PAMAP2  & 4        & \begin{tabular}[c]{@{}c@{}}8\\ (No.1,2,3,4,\\ 12,13,16,17)\end{tabular} & 8               & \begin{tabular}[c]{@{}c@{}}acc($\pm16g$),\\  gyro,\\  mag\end{tabular} & 512           & 5.12s           & (27, 1, 512)  & (1, 9)      & \begin{tabular}[c]{@{}c@{}}2 conv2d layers\\ (Relu-MaxPool2d)\\ 1 fc layer\\ (Relu)\end{tabular} \\\hline 
USC-HAD & 5        & 14                                                                      & 12              & \begin{tabular}[c]{@{}c@{}}acc,\\  gyro\end{tabular}           & 500           & 5s              & (6, 1, 500)  & (1, 6)      & \begin{tabular}[c]{@{}c@{}}3 conv2d layers\\ (Relu-MaxPool2d)\\ 1 fc layer\\ (Relu)\end{tabular} \\ \bottomrule
\end{tabular}}

\end{table*}



\end{document}